%% file: main.tex
\documentclass[10pt,twocolumn,letterpaper]{article}

\usepackage[pagenumbers]{cvpr}
\input{preamble}

\definecolor{cvprblue}{rgb}{0.21,0.49,0.74}
\usepackage[pagebackref,breaklinks,colorlinks,allcolors=cvprblue]{hyperref}

\def\confName{CVPR}
\def\confYear{2027}

\title{RoboRecast: Intent-Preserving Demonstration Augmentation via Diagnosis-Driven Optimization}

\author{Yushi Long$^{1}$ \quad Juzhan Xu$^{1}$ \quad Haibin Huang$^{3}$ \quad Chi Zhang$^{3}$\\
Yin Chen$^{3}$ \quad Sida Peng$^{1}$ \quad Ruizhen Hu$^{2,\dagger}$ \quad Xiaowei Zhou$^{1}$\\[3pt]
$^{1}$Zhejiang University \qquad $^{2}$Shenzhen University \qquad $^{3}$China Telecom}

\begin{document}

\makeatletter
\apptocmd{\@maketitle}{\input{floats/teaser}}{}{}
\makeatother
\maketitle
\begingroup
\renewcommand{\thefootnote}{\ensuremath{\dagger}}
\footnotetext{Corresponding author.}
\endgroup
\addtocontents{toc}{\protect\setcounter{tocdepth}{-1}}
\input{sec/0_abstract}    
\input{sec/1_intro}

\input{floats/overview}
\input{sec/2_related_work}

\input{sec/3_method}

\input{sec/4_experiments}

\input{sec/5_conclusion}
\input{sec/6_llm_usage_statement}

{
    \small
    \bibliographystyle{ieeenat_fullname}
    \bibliography{main}
}

\clearpage
\maketitlesupplementary
\appendix
\setcounter{figure}{0}
\setcounter{table}{0}
\renewcommand{\thefigure}{S\arabic{figure}}
\renewcommand{\thetable}{S\arabic{table}}
\renewcommand{\theHfigure}{supp.\arabic{figure}}
\renewcommand{\theHtable}{supp.\arabic{table}}
\addtocontents{toc}{\protect\setcounter{tocdepth}{2}}
\input{sec/appendix/contents}

\input{sec/appendix/a_implementation}

\input{sec/appendix/b_evaluation_protocols_and_additional_results}

\input{sec/appendix/c_limitations_and_future_work}
\end{document}

%% file: preamble.tex
\usepackage{multirow}

%% file: floats/teaser.tex
\begin{minipage}{\textwidth}
  \centering
  \includegraphics[width=\linewidth]{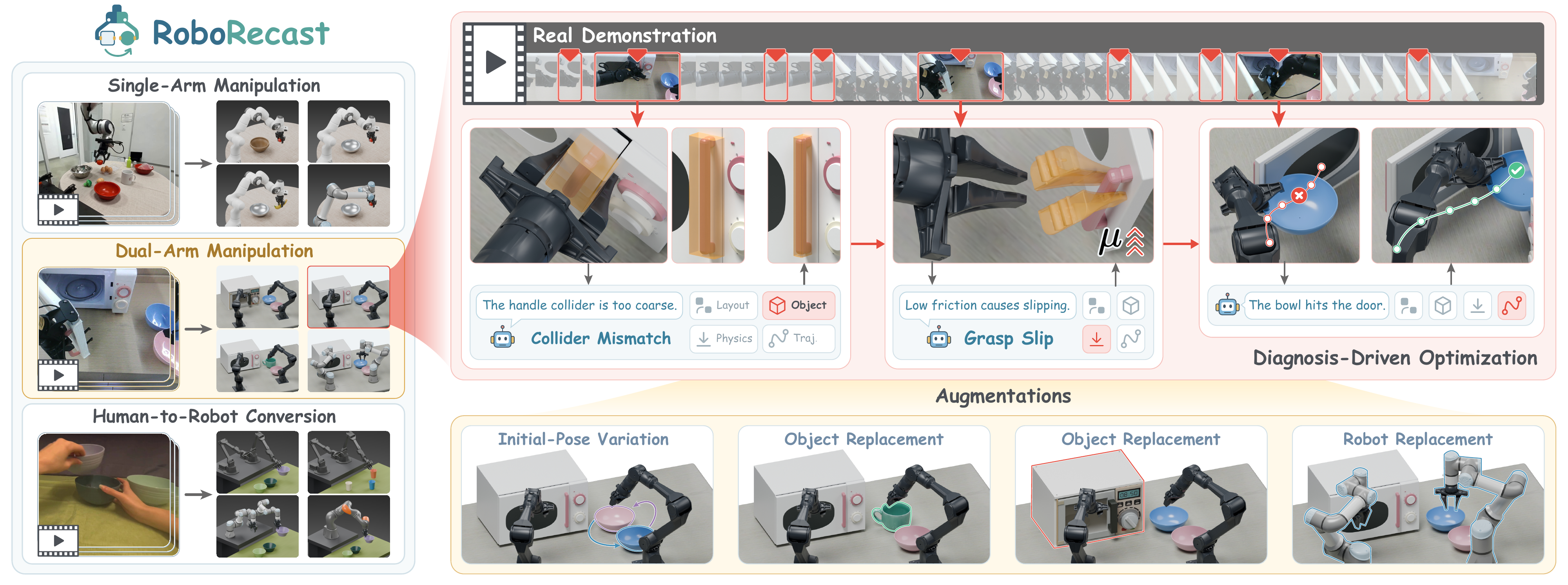}
  \captionof{figure}{\textbf{RoboRecast} augments real demonstrations by preserving their interaction requirements while adapting scenes and trajectories for physically plausible execution. The bimanual example schematically illustrates how simulation feedback guides corrections to collision geometry, friction, and waypoints. The framework supports augmentation across demonstration types—including single-arm robot, bimanual robot, and human demonstrations—through initial-pose variation, object replacement, and robot replacement.}
  \label{fig:teaser}
\end{minipage}
\vspace{1em}

%% file: sec/0_abstract.tex
\begin{abstract}
Collecting diverse real-world manipulation demonstrations is costly, motivating the use of simulation to augment existing demonstrations for robot policy learning.
When objects, layouts, or robots change, transferring motions from a source demonstration can produce behavior that is semantically implausible or physically infeasible in the new setting.
We introduce RoboRecast, an agentic optimization framework that preserves the interactions required by the source task while jointly optimizing the simulated scene and robot motion to make those interactions executable in the target setting.
When an interaction fails, the agent uses simulation feedback to diagnose the cause, revise the relevant scene or motion parameters, and test whether the revision resolves the failure.
This flexibility allows RoboRecast to generate synthetic data from both human and robot demonstrations and to adapt them to new layouts, objects, and robot embodiments.
When target demonstrations are available, adding RoboRecast data improves policy success on both real-world single-arm and simulated bimanual tasks.
Even without demonstrations in the target configuration, RoboRecast adapts source demonstrations to support policy learning with new objects or robot embodiments.
\par\noindent{\small\normalfont Project Page: \href{https://zju3dv.github.io/RoboRecast/}{zju3dv.github.io/RoboRecast/}}\par
\vspace{-\baselineskip}
\end{abstract}

%% file: sec/1_intro.tex
\section{Introduction}
\label{sec:intro}

Robot policies require diverse demonstrations to generalize across objects, layouts, and robot embodiments, yet collecting such coverage in the real world is costly, motivating the augmentation of existing demonstrations.
For the resulting synthetic data to support policy learning, the generated interactions must preserve the source manipulation intent and remain physically executable with the specified objects, layout, and robot.
Image and video augmentation typically enriches observations around given robot motions~\cite{chen2023genaug,yu2023scaling,wang2026robovip,liu2026robotransfer,ye2026robodream}, limiting its ability to accommodate changes in initial object poses, object assets, or robot embodiments.
Recreating interactions in simulation allows scenes and motions to be varied together, with explicit control and execution feedback for checking and improving physical plausibility.
Such control and execution-based refinement make simulation a promising route to augmenting real-world manipulation demonstrations.

The goal of simulation-based augmentation is to generate synthetic data in which the robot completes the demonstrated task under the physical constraints of the target setting.
Existing work provides two building blocks for this goal: real-to-sim methods reconstruct the environment~\cite{zhang2026robosnap,han2026re,chen2026agentic}, while demonstration transfer adapts robot motions to new configurations.
Errors in reconstructed geometry or physical properties can prevent the intended interaction, and adjusting the motion alone leaves those errors unresolved.
On the motion side, demonstration transfer adapts source motion segments~\cite{mandlekar2023mimicgen,garrett2024skillmimicgen,jiang2025dexmimicgen} or keypoint-trajectory constraints~\cite{lin2025constraint} to the target configuration.
When a new object or layout calls for a different grasp or approach, these inherited constraints can rule out feasible motions or produce unnatural movements.
These limitations raise two questions: what must be preserved from the source demonstration, and how should the scene and robot motion be adapted to the target configuration?

Our key insight is that the same task intent can be realized in different ways, so augmentation should preserve the intended effects of the interaction while allowing the scene and motion to adapt.
We therefore propose expressing the intended effects as interaction requirements for each phase and jointly optimizing the scene and robot motion to satisfy them under observation, target, and physical constraints.
Inferring these requirements from a real-world demonstration requires understanding what the demonstrator is doing and what each phase is intended to achieve.
Even with these requirements identified, the search spans a large space of object geometries, physical properties, and robot motions, while evaluating a candidate requires simulation rollouts that provide no direct gradient for improving it.

We introduce \textbf{RoboRecast} (Fig.~\ref{fig:teaser}), an agentic optimization framework that uses an agent to infer interaction requirements from demonstrations and uses simulation feedback to guide joint optimization of scenes and robot motions.
An agent is well suited to this problem because its semantic understanding and reasoning enable it to infer the intent behind observed actions, diagnose likely causes of execution failures, and revise the scene or motion accordingly.
By interpreting object roles, contacts, and resulting state changes in the source demonstration, the agent identifies the essential interactions and their ordering, then maps these requirements to the target configuration.
When a phase fails to meet one of these fixed requirements or violates a physical constraint, the agent examines visual and numerical evidence from that phase to identify a likely cause in the reconstructed scene or robot motion.
This diagnosis narrows each update to the relevant object parameters or motion waypoints, allowing the agent to test focused revisions in simulation and reduce trial and error.

With target demonstrations available, RoboRecast reconstructions yield 38.3\% policy success across three real-to-sim tasks, compared with 18.3\% for RoboSnap~\cite{zhang2026robosnap} and 23.3\% for AR2S~\cite{chen2026agentic}.
Across four layout augmentation tasks, RoboRecast data yields 53.8\% policy success, compared with 33.8\% for MimicGen~\cite{mandlekar2023mimicgen} and 40.0\% for CP-Gen~\cite{lin2025constraint}.
On the same four tasks, adding robot trajectories converted from human demonstrations yields 46.3\% policy success with RoboRecast and 32.5\% with Ego2Robot~\cite{wang2026ego2robot}.
Our synthetic data also improves policy success in bimanual simulation.
Even without demonstrations in the target configuration, RoboRecast adapts source demonstrations to support policy learning with new objects or robots.

Our contributions are threefold:
\begin{itemize}
\setlength{\itemsep}{2pt}
\setlength{\parsep}{0pt}
\item We formulate demonstration augmentation as joint scene and motion optimization, preserving task-defining interaction requirements while allowing their realization to adapt to target configurations.
\item We introduce RoboRecast, an agentic optimization framework that extracts interaction requirements from demonstrations and jointly optimizes scenes and motions through execution-guided diagnosis and targeted updates.
\item We demonstrate that RoboRecast generates effective policy training data both with and without demonstrations available for the target configuration.
\end{itemize}

%% file: floats/overview.tex
\begin{figure*}[t]
  \centering
  \includegraphics[width=\textwidth]{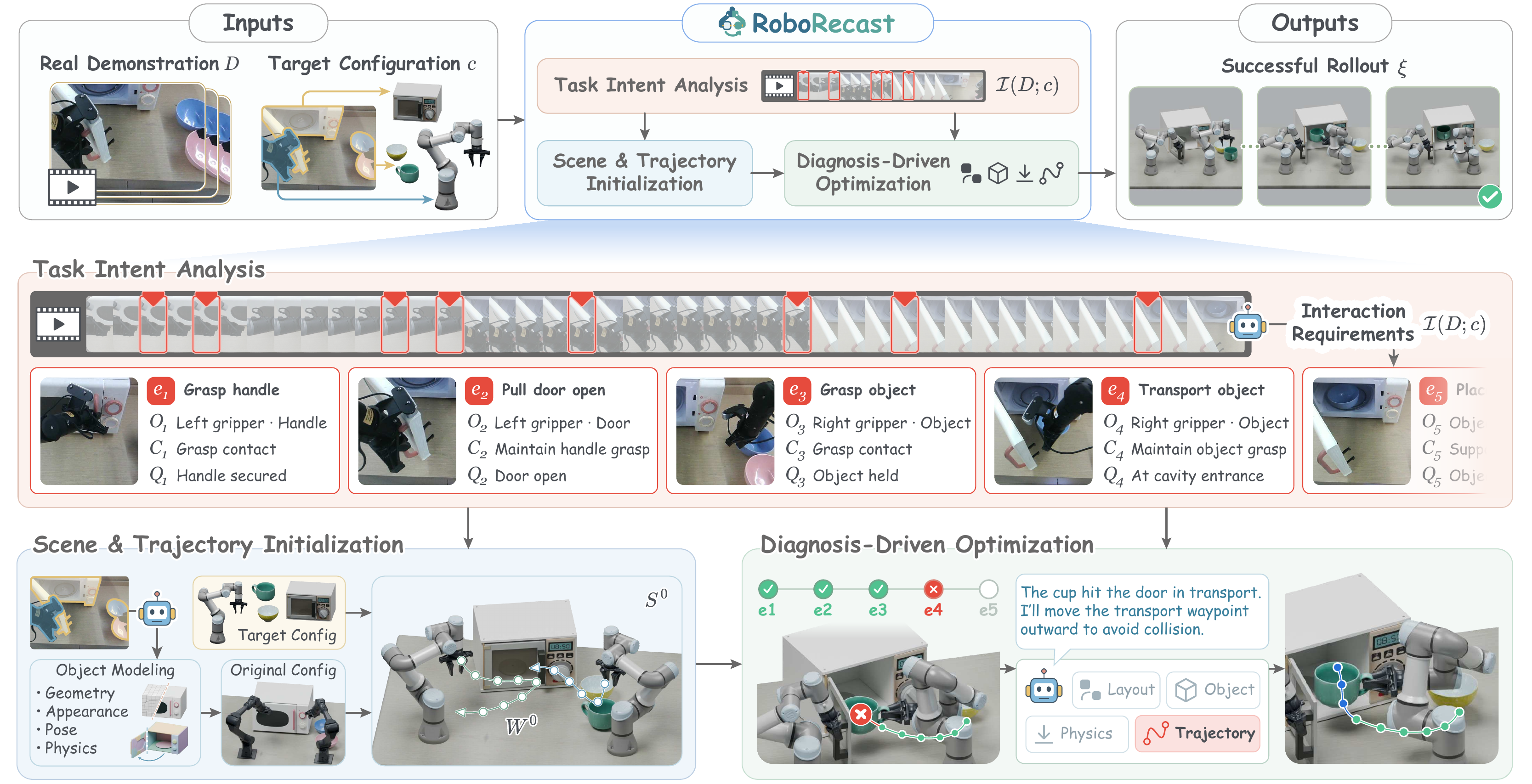}
  \caption{\textbf{Overview of RoboRecast.}
  Given the real demonstration and target configuration, RoboRecast generates a successful simulation rollout that preserves the demonstrated task intent under that configuration.
  The agent first extracts the interaction requirements that define the task intent and uses them to initialize the target scene and robot trajectory.
  When a validity check or simulation reveals a failure, the agent uses visual and numerical feedback to diagnose its cause, revise the relevant scene or trajectory parameters, and test the revision.}
  \label{fig:overview}
\end{figure*}

%% file: sec/2_related_work.tex
\section{Related Work}
\label{sec:related_work}

\noindent\textbf{Image and Video Augmentation.}
Image editing and video synthesis expand the range of visual conditions covered by robot demonstrations.
GenAug~\cite{chen2023genaug} and ROSIE~\cite{yu2023scaling} edit demonstration images to vary objects, backgrounds, and distractors.
RoboVIP~\cite{wang2026robovip} and RoboTransfer~\cite{liu2026robotransfer} emphasize temporally consistent multi-view video generation, while RoboDream~\cite{ye2026robodream} conditions synthesis on robot motion, object priors, and scene priors.
Because these methods primarily vary observations around a given motion, they offer limited flexibility when changes in initial layout, object, or robot embodiment require a new grasp or path.
Moreover, their appearance and geometry controls do not establish physical executability.

\noindent\textbf{Real-to-Sim for Manipulation Data Generation.}
Real-to-sim methods turn real-world observations into interactive environments for demonstration generation and policy learning\cite{heppert2026scaling}.
RE$^3$Sim~\cite{han2026re} combines multi-view reconstruction with neural rendering, while RoboSnap~\cite{zhang2026robosnap} builds interactive foreground assets and a Gaussian-splatting background from a single image.
RE$^3$Sim generates motions using scripts that access object poses, while ReBot~\cite{fang2025rebot} replays recorded robot trajectories.
AR2S~\cite{chen2026agentic} uses vision-language models to support scene reconstruction and refines replayed interactions through simulator feedback, including object-placement adjustments for grasping.
Even after such adjustments, errors in reconstructed geometry can prevent the intended contact.
RoboRecast therefore uses execution feedback to choose between scene and motion updates.

\noindent\textbf{Demonstration Transfer and Augmentation.}
Demonstration transfer adapts source motions to new configurations, reducing the need for additional demonstrations~\cite{xue2025demogen,ameperosa2025rocoda}.
MimicGen~\cite{mandlekar2023mimicgen} transforms object-relative segments, SkillGen~\cite{garrett2024skillmimicgen} connects manipulation skills through planning, and DexMimicGen~\cite{jiang2025dexmimicgen} extends transfer to bimanual and dexterous manipulation.
CP-Gen~\cite{lin2025constraint} adapts keypoint-trajectory constraints to object variations, accounts for smoothness during trajectory optimization, and uses motion planning to connect segments.
When a target configuration calls for a different grasp or approach direction, transferred motion segments or adapted keypoint constraints may exclude feasible motions or produce unnatural movements.
We instead fix interaction requirements---object roles, essential contacts, and state changes---while allowing paths, timing, and unconstrained contact choices to adapt.

\noindent\textbf{Agentic 3D Modeling and Manipulation.}
Agentic methods coordinate specialized tools and feedback for 3D modeling and robot manipulation.
For 3D asset modeling, AgentSTAR~\cite{mazur2026agentstar} reconstructs and tracks objects from video, while Articraft~\cite{zhou2026articraft} generates articulated assets.
At the scene level, Lucida~\cite{qin2026lucida} reconstructs observed scenes, while SceneSmith~\cite{pfaff2026scenesmith} and SAGE~\cite{xia2026sage} generate environments from language or task descriptions.
In concurrent work, ARSTAG~\cite{li2026arstag} coordinates reconstruction and data generation through cross-stage feedback, while Real2Gym~\cite{ren2026real2gym} builds interactive environments from demonstrations for skill acquisition and transfer.
Our agent extracts interaction requirements from the source demonstration and uses them to diagnose which scene parameters or semantic waypoints need revision.

%% file: sec/3_method.tex
\section{Method}
\label{sec:method}

\noindent\textbf{Overview.}
Given a real demonstration $D$ and a target configuration $c$ (objects, layout, and robot), RoboRecast seeks a successful rollout $\xi$ that preserves the demonstrated task intent (Fig.~\ref{fig:overview}).
The agent infers the required interactions from $D$, maps their roles to target entities, and initializes the scene and trajectory.
If a validity check fails or the rollout misses a required interaction, the agent uses the evidence to diagnose the failure and update the scene or motion.

\subsection{Task Intent Analysis and Problem Formulation}
\label{sec:task_formulation}

\noindent\textbf{Interaction Requirement Extraction.}
To infer what each interaction must accomplish, the agent examines keyframes and their surrounding segments in the source demonstration $D$ to identify the participating roles, essential contacts, and intended state changes.
The agent then maps each participating role to the corresponding entity in the target configuration $c$.
We represent the task intent under $c$ as a partially ordered set of interaction requirements,
\begin{equation}
\mathcal I(D;c)=\left(\{e_i\}_{i=1}^{K},\prec\right),\quad
 e_i=(\mathcal O_i(c),\mathcal C_i,\mathcal Q_i),
\label{eq:interaction_requirements}
\end{equation}
where $\mathcal O_i(c)$ identifies the target entities and $\mathcal C_i$ and $\mathcal Q_i$ specify the required contacts and state changes among those entities.
The agent infers the precedence relation $\prec$ from the demonstration, retaining the ordering needed for the task while allowing independent bimanual actions to overlap.
With these requirements fixed, the optimizer can adjust the scene and motion to realize them in the target configuration.

\noindent\textbf{Scene and Trajectory Parameterization.}
We represent the target scene before manipulation as
\begin{equation}
\begin{aligned}
S&=(G,\Theta),\\
\Theta&=\bigl\{\theta_n^{\mathrm{pose}},\theta_n^{\mathrm{shape}},\theta_n^{\mathrm{app}},\theta_n^{\mathrm{col}},
\theta_n^{\mathrm{phys}},\theta_n^{\mathrm{joint}}\bigr\}_{n=1}^{N_o}.
\end{aligned}
\label{eq:scene_representation}
\end{equation}
The scene graph $G$ records relationships among objects.
For each of the $N_o$ objects, $\Theta$ specifies its pose, visible geometry, appearance, collision geometry, physical properties, and articulation.
During optimization, the scene graph $G$ remains fixed, while object attributes in $\Theta$ are editable.

Within this scene, we parameterize the robot motion with sparse semantic waypoints $W=\{w_j\}_{j=1}^{M}$, where
\begin{equation}
w_j=\bigl(p_j,R_j,g_j,\Delta t_j,\ell_j,F_j(S)\bigr),\quad
 t_j=\sum_{q=1}^{j}\Delta t_q.
\label{eq:semantic_waypoint}
\end{equation}
Each waypoint specifies an end-effector position $p_j\in\mathbb R^3$ and orientation $R_j\in\mathrm{SO}(3)$ relative to a reference frame $F_j(S)$, which may be the world frame or a frame attached to an object or interaction region in $S$.
The gripper command $g_j$ specifies the gripper action, and $\Delta t_j>0$ gives the time elapsed since the preceding waypoint, with $t_0=0$.
The label $\ell_j$ specifies the local motion mode: \textit{transit}, \textit{dwell}, or \textit{sustained contact}.
For bimanual tasks, each arm has its own waypoint sequence on a shared timeline.
For each interaction requirement $e_i$, the waypoints intended to realize it delimit a time interval $\mathcal T_i$ over which the resulting interaction is evaluated.

\noindent\textbf{Joint Optimization Formulation.}
We jointly optimize the target scene $S$ and the semantic waypoints $W$ to satisfy the interaction requirements $\mathcal I(D;c)$.
For each candidate $(S,W)$, we compile the waypoints into robot commands and execute them in $S$ to obtain the rollout
\begin{equation}
\xi(S,W)=\operatorname{Sim}\bigl(S,\operatorname{Compile}(W;S)\bigr).
\label{eq:demonstration_generation}
\end{equation}
The compiler uses $S$ to express the waypoint poses in a common frame, then interpolates between them and applies inverse kinematics for the target robot (Sec.~\ref{sec:diagnosis_refinement}).
For each requirement $e_i$, the agent examines the rollout over $\mathcal T_i$ for the required contacts and state changes and uses the rollout timing to verify the precedence relations involving $e_i$.
We record this assessment with a binary indicator,
\begin{equation}
b_i(S,W)=
\begin{cases}
1, & \text{if } \xi(S,W) \text{ satisfies } e_i,\\
0, & \text{otherwise}.
\end{cases}
\label{eq:interaction_satisfaction}
\end{equation}
We therefore minimize the number of unmet interaction requirements subject to physical, observation, and target constraints,
\begin{equation}
\begin{aligned}
\min_{S,W}\quad &\sum_{i=1}^{K}\bigl[1-b_i(S,W)\bigr]\\
\mathrm{s.t.}\quad &C_{\mathrm{phys}}(S,\xi)=\mathrm{true},\\
 &C_{\mathrm{obs,target}}(S;D,c)=\mathrm{true}.
\end{aligned}
\label{eq:augmentation_constraints}
\end{equation}
The physical constraint $C_{\mathrm{phys}}$ requires scene stability, physically valid contacts, and robot motion within joint and velocity limits, ruling out excessive penetration and unintended collisions.
The constraint $C_{\mathrm{obs,target}}$ requires the scene to reflect the changes specified by $c$ while remaining consistent with the source observations in all other respects.
RoboRecast accepts a rollout only when the objective is zero and both constraints hold.

\subsection{Scene and Trajectory Initialization}
\label{sec:initialization}
To initialize the joint optimization in Eq.~\ref{eq:augmentation_constraints}, the agent constructs a scene for the target configuration $c$ and a robot motion for the interaction requirements $\mathcal I(D;c)$, yielding the initial candidate $(S^0,W^0)$.

\noindent\textbf{Scene Initialization.}
From the demonstration $D$, the agent builds the source scene graph $G_{\mathrm{src}}$ by identifying the interacting objects, their parts, and their spatial relationships.
SAM~3~\cite{carion2026sam} segments objects in keyframes, and LingBot-Depth~\cite{tan2026masked} refines sensor depth.
Using these observations, the agent follows the iterative workflow from AgentSTAR~\cite{mazur2026agentstar} to reconstruct each object's geometry $\theta_n^{\mathrm{shape}}$ and appearance $\theta_n^{\mathrm{app}}$ and estimate its pose $\theta_n^{\mathrm{pose}}$.

For each reconstructed object, the agent generates collision geometry $\theta_n^{\mathrm{col}}$ by convex decomposition at adjustable granularity~\cite{shi2026cuacd}.
Where overlays on the visual mesh reveal missing detail, the agent refines the collision geometry with scripted box primitives.
The agent models the joint structure $\theta_n^{\mathrm{joint}}$ where needed and initializes the physical properties $\theta_n^{\mathrm{phys}}$ using density and friction values for the inferred material.
Using $G_{\mathrm{src}}$, the agent assembles the source scene and adjusts object poses to resolve interpenetration and restore observed support contacts.

The agent then applies $c$ to reposition objects, replace an object, or switch robots, updating the affected relationships in $G_{\mathrm{src}}$ to obtain the target graph $G$.
For a replacement object, the agent reuses the simulation properties of a simulation-ready asset or builds them with the same procedure if only a visual mesh is available.
Together, $G$ and the initialized object attributes $\Theta^0$ define the target scene $S^0=(G,\Theta^0)$.

\noindent\textbf{Trajectory Initialization.}
Within the initial target scene $S^0$, the agent constructs the semantic waypoint sequence $W^0$ for the target robot from the mapped interaction requirements.
For each requirement $e_i$, the agent chooses a contact region and approach direction based on the target geometry, then places waypoints to establish or maintain the required contact.
For bimanual tasks, the agent assigns roles to the two arms and coordinates their timing according to the required order of interactions.
As it builds $W^0$, the agent records which waypoint segment corresponds to each requirement $e_i$, so it can evaluate that interaction over the corresponding rollout interval $\mathcal T_i$.

\subsection{Diagnosis-Driven Optimization}
\label{sec:diagnosis_refinement}
The agent runs all scene and motion validity checks on the initial candidate $(S^0,W^0)$ and repeats only the affected checks after each edit.
Once the required checks pass, the agent simulates the motion to verify the interaction requirements phase by phase.
If a check fails or the rollout reveals an unmet requirement or physical violation, the agent diagnoses the likely cause and tests a targeted update to the relevant scene parameters or waypoints (Table~\ref{tab:variable_blocks}).

\noindent\textbf{Validity Checks and Trajectory Execution.}
Scene validity checks assess both the consistency of the candidate scene $S$ with the intended setting and its physical plausibility.
The scene should reflect the changes specified by the target configuration $c$, while rendered views of retained elements should match observations in the source demonstration $D$.
To assess physical plausibility, the tools compare collision geometry with object surfaces and test whether objects settle stably and articulated parts move as intended.

To compile the motion, the compiler resolves each waypoint frame $F_j(S)$ in the current scene and interpolates between the resulting world-frame poses.
Inverse kinematics converts the interpolated poses into joint commands, while the gripper follows the timing specified by the waypoints.
Motion validity checks flag target poses for which no IK solution is found and joint trajectories that exceed position or velocity limits.
Once the required checks pass, the tools execute the motion in MuJoCo~\cite{todorov2012mujoco} and record the rollout $\xi$, including rendered keyframes and numerical records of object motion and contact.

\noindent\textbf{Failure Diagnosis.}
When a validity check fails, the agent uses the reported violation to identify the scene element or motion segment to investigate.
For an executed trajectory, the agent evaluates each requirement $e_i$ over its interval $\mathcal T_i$ and investigates the earliest unmet requirement or physical violation in the rollout.
In Fig.~\ref{fig:overview}, comparing keyframes from the transport phase of the source demonstration and the rollout shows that the simulated cup fails to clear the door.
The contact log confirms that the cup strikes the door during transport.
If the door is correctly positioned and has opened as required, the agent suspects insufficient clearance along the transport path and proposes moving the waypoint outward for the next rollout.

\begin{table}[t]
\centering
\small
\caption{Editable scene and trajectory parameters (Eqs.~\ref{eq:scene_representation} and~\ref{eq:semantic_waypoint}).}
\label{tab:variable_blocks}
\begin{tabular}{@{}ll@{}}
\toprule
Parameter category & Editable parameters \\
\midrule
Scene Configuration & $\theta_n^{\mathrm{pose}}$ \\
Object Structure & $\theta_n^{\mathrm{shape}},\theta_n^{\mathrm{col}},\theta_n^{\mathrm{joint}}$ \\
Physical Properties & $\theta_n^{\mathrm{phys}}$ \\
Trajectory & $W=\{w_j\}_{j=1}^{M}$ \\
\bottomrule
\end{tabular}
\end{table}

\noindent\textbf{Targeted Updates and Verification.}
Based on its diagnosis, the agent specifies which object attribute or waypoint to change and how to change it.
The editing tool applies the requested change to the object in the scene or modifies the waypoint sequence for the selected phase, including inserting a waypoint when needed.
The agent associates any new waypoint with its interaction requirement and updates the affected evaluation intervals $\mathcal T_i$ if the timing changes.
The tools recompile the motion as needed, repeat the affected validity checks, and rerun the affected motion once those checks pass.
From the resulting evidence, the agent determines whether the edit resolved the original failure and whether it introduced any new violations.

\noindent\textbf{Checkpoint Reuse.}
After verifying the interactions in a phase, RoboRecast saves the simulation state at the end of that phase as a checkpoint.
After a local edit, execution resumes from the latest checkpoint whose saved state is compatible with the edit and whose earlier interactions remain verified.
If no checkpoint meets both conditions, execution restarts from the beginning.
When a checkpoint is reused, the agent retains the verification results for the preceding phases and reviews only evidence relevant to the edit.

%% file: sec/4_experiments.tex
\section{Experiments}
\label{sec:experiments}

With target demonstrations, we test whether real-to-sim reconstruction, layout augmentation, and human-to-robot conversion improve policy success on a physical UR5, and whether cross-simulator augmentation improves bimanual success on RoboTwin (Sec.~\ref{sec:policy_learning}).
Without target demonstrations, we evaluate real-world policies trained solely on trajectories adapted to a new object or robot (Sec.~\ref{sec:cross_configuration_transfer}).
Ablations measure how interaction requirements and diagnosis-driven updates affect generation success and optimization time (Sec.~\ref{sec:ablation_studies}).

\subsection{Experimental Setup}
\label{sec:experimental_setup}
\noindent\textbf{Platforms.} We evaluate single-arm policies on a physical UR5 and use a physical Piper-X for robot adaptation.
For bimanual evaluation, we use RoboTwin's default Aloha configuration~\cite{chen2025robotwin} in a cross-simulator setting spanning SAPIEN~\cite{xiang2020sapien} and MuJoCo~\cite{todorov2012mujoco}.

\noindent\textbf{Policy Training.} We train all policies using Diffusion Policy~\cite{chi2025diffusion} with external-camera images as visual input.
At each step of a successful RoboRecast rollout, we export the observations the policy requires and the corresponding action.
We compare methods by adding equal numbers of trajectories per task to a shared base dataset while holding policy initialization, training steps, and hyperparameters fixed.

\noindent\textbf{Evaluation.} We report full-task success over 20 real-world trials or 100 RoboTwin trials per task.
For real-world tasks, we also report stage success, such as grasping the object or pushing it partway to the goal.

\subsection{Augmentation with Target Demonstrations}
\label{sec:policy_learning}
\input{floats/real_world_setup}
\input{floats/real_policy}

We collected 50 demonstrations per task by teleoperating a UR5 (Real-50).
Across four tasks---stacking cubes, placing a cube into a bowl, pushing a cube across a line, and opening a microwave (Fig.~\ref{fig:real_world_setup})---we compare three ways to supplement Real-50: real-to-sim reconstruction, layout augmentation, and human-to-robot conversion (Table~\ref{tab:real_policy}).

\input{floats/ur5_reconstruction}
\noindent\textbf{Real-to-sim reconstruction.} We first supplement Real-50 with 50 trajectories per task reconstructed by RoboSnap~\cite{zhang2026robosnap}, AR2S~\cite{chen2026agentic}, or RoboRecast.
Adding RoboRecast reconstructions to Real-50 raises mean full-task success from 22.5\% to 36.3\% across all four tasks, with success on the three rigid-object tasks reaching 38.3\% for RoboRecast versus 18.3\% for RoboSnap and 23.3\% for AR2S (Table~\ref{tab:real_policy}).
Figure~\ref{fig:ur5_reconstruction} suggests one reason for this gap.
RoboRecast more closely matches the relative sizes of the cube and bowl and the distance from the cube to the target line, so its demonstrations better reflect the placement tolerance and pushing distance required on the real robot.
For microwave opening, AR2S cannot model the articulated microwave, and RoboSnap's open-source pipeline requires manual assistance.
The appendix reports additional reconstruction results on the initial frames of DROID\cite{khazatsky2024droid} demonstrations.

\noindent\textbf{Layout augmentation.} Given the same 50 RoboRecast reconstructions per task, MimicGen~\cite{mandlekar2023mimicgen}, CP-Gen~\cite{lin2025constraint}, and RoboRecast each generate 50 trajectories with varied initial object positions.
We then add each method's augmented trajectories to the same training set of 50 real demonstrations and 50 RoboRecast reconstructions.
Mean full-task success reaches 33.8\% with MimicGen, 40.0\% with CP-Gen, and 53.8\% with RoboRecast, up from 36.3\% before augmentation (Table~\ref{tab:real_policy}).
MimicGen transfers object-relative motion segments, while CP-Gen adapts keypoint trajectory constraints from the source demonstration.
RoboRecast preserves the required contacts and state changes while allowing the robot to approach and move the object along a path suited to its new position.
Across both comparisons, gains in full-task success exceed those in stage success, consistent with more natural and coherent motions whose training benefits may accumulate across successive interaction stages.

\noindent\textbf{Human-to-robot conversion.} We collected 50 human demonstrations per task and converted them into robot trajectories using either Ego2Robot~\cite{wang2026ego2robot} or RoboRecast.
Adding these trajectories to Real-50 raises mean full-task success from 22.5\% to 32.5\% with Ego2Robot and to 46.3\% with RoboRecast (Table~\ref{tab:real_policy}).
Ego2Robot retargets the observed hand motion to the robot and synthesizes the corresponding images.
RoboRecast further tests whether the gripper's contact produces the intended object motion in simulation, and revises the interaction when it fails.
Adding 50 layout variations of the converted trajectories per task (OurH2RAug-50) further raises mean success to 56.3\%.

\input{floats/bimanual_policy}
\noindent\textbf{Bimanual cross-simulator evaluation.} Across four bimanual RoboTwin tasks, adding 50 of our cross-simulator reconstructions to the 50 original demonstrations per task raises mean task success from 36.5\% to 39.8\% (Table~\ref{tab:bimanual_policy}).
Adding a further 50 layout augmentations per task to this shared training set brings mean success to 46.3\% with RoboRecast and 41.5\% with DexMimicGen~\cite{jiang2025dexmimicgen}.
The gains suggest that RoboRecast's layout variations also support policy learning when the arms must coordinate their contacts, such as in microphone handover, where one gripper must secure the microphone before the other releases it.

\subsection{Augmentation without Target Demonstrations}
\label{sec:cross_configuration_transfer}
We next consider tasks involving a new object or robot for which no target demonstrations are available.
Given only demonstrations from the source configuration, can RoboRecast generate trajectories that suffice to train a policy for the target task on a real robot?

\noindent\textbf{Target-object adaptation.} We evaluate the UR5 on stacking a mouse, placing a Joy-Con in a bowl, and pushing a mug across a line (Fig.~\ref{fig:ur5_object_adaptation}), using only cube manipulation demonstrations as source data.
On the target objects, policies trained on 50 original demonstrations per task achieve 1.7\% mean full-task success, compared with 21.7\% for policies trained on 50 trajectories synthesized for those objects (Table~\ref{tab:object_adaptation}).
Increasing the number of synthesized trajectories to 100 per task raises mean success to 33.3\%.

\input{floats/policy_results_layout}
\input{floats/ur5_to_piper_x}
\noindent\textbf{Target-robot adaptation.} We evaluate policies on a physical Piper-X across four tasks for which only UR5 demonstrations are available.
RoboRecast adapts these interactions to Piper-X in simulation (Fig.~\ref{fig:ur5_to_piper_x}).
Policies trained solely on these trajectories improve on all four tasks as the training set grows from 50 to 100 trajectories per task, raising mean full-task success from 12.5\% to 22.5\% (Table~\ref{tab:robot_adaptation}).

\subsection{Ablation Studies}
\label{sec:ablation_studies}

We examine how explicit interaction requirements and diagnosis-driven updates to the scene and motion affect RoboRecast's ability to generate successful rollouts efficiently.
For each of 20 DROID~\cite{khazatsky2024droid} demonstrations, all variants start from the same initialized scene and waypoints.
Each optimization run allows up to ten rollouts and stops early if the agent reports that the task is complete.
We report mean optimization time across demonstrations, measured from the end of initialization.

\noindent\textbf{Explicit interaction requirements.} We test whether the agent benefits from explicit requirements for each stage by comparing against a variant prompted only to reproduce the source trajectory and correct its execution.
Both variants use the same tools, editable parameters, and rollout budget, and we manually assess whether each final rollout completes the task.
With explicit requirements, 20 of 20 rollouts succeed (100\%) in 223 seconds on average, compared with 13 of 20 (65\%) in 348 seconds for trajectory reproduction.
These results suggest that specifying the intended effect of each stage helps the agent identify where execution first falls short and focus its edits on that interaction.

\noindent\textbf{Diagnosis-driven joint optimization.} Once the agent identifies a failed interaction, it must determine whether the cause lies in the scene or the motion and which parameters to revise.
We test how the choice of editable parameters and the scope of each update affect optimization (Table~\ref{tab:trajectory_ablation}).
From the same initialization, task success is 30\% without optimization and 75\% when the agent can revise only the trajectory.
When both scene and trajectory parameters are editable, allowing the agent to revise multiple blocks per update yields 90\% success in 388 seconds on average.
Targeting one block at a time based on the diagnosed failure yields 100\% success and reduces mean optimization time to 223 seconds.
This suggests that some failures require scene corrections and that targeted edits can reduce the risk of introducing new errors while resolving a failure.

\input{floats/trajectory_ablation}
\noindent\textbf{Execution efficiency and motion quality.} Removing either checkpointing or waypoint interpolation preserves success on all 20 demonstrations (Table~\ref{tab:trajectory_ablation}).
Without checkpointing, the simulator replays previously verified phases even when an edit leaves them unaffected, and the agent may recheck their requirements, increasing mean optimization time from 223 to 315 seconds.
Without interpolation, the agent must specify more waypoints, raising mean optimization time to 280 seconds and producing less smooth motion at both the joints and the end effector, as measured by J-LDLJ and E-LDLJ defined in the appendix.

%% file: floats/real_world_setup.tex
\begin{figure}[tbp]
  \centering
  \includegraphics[width=\columnwidth]{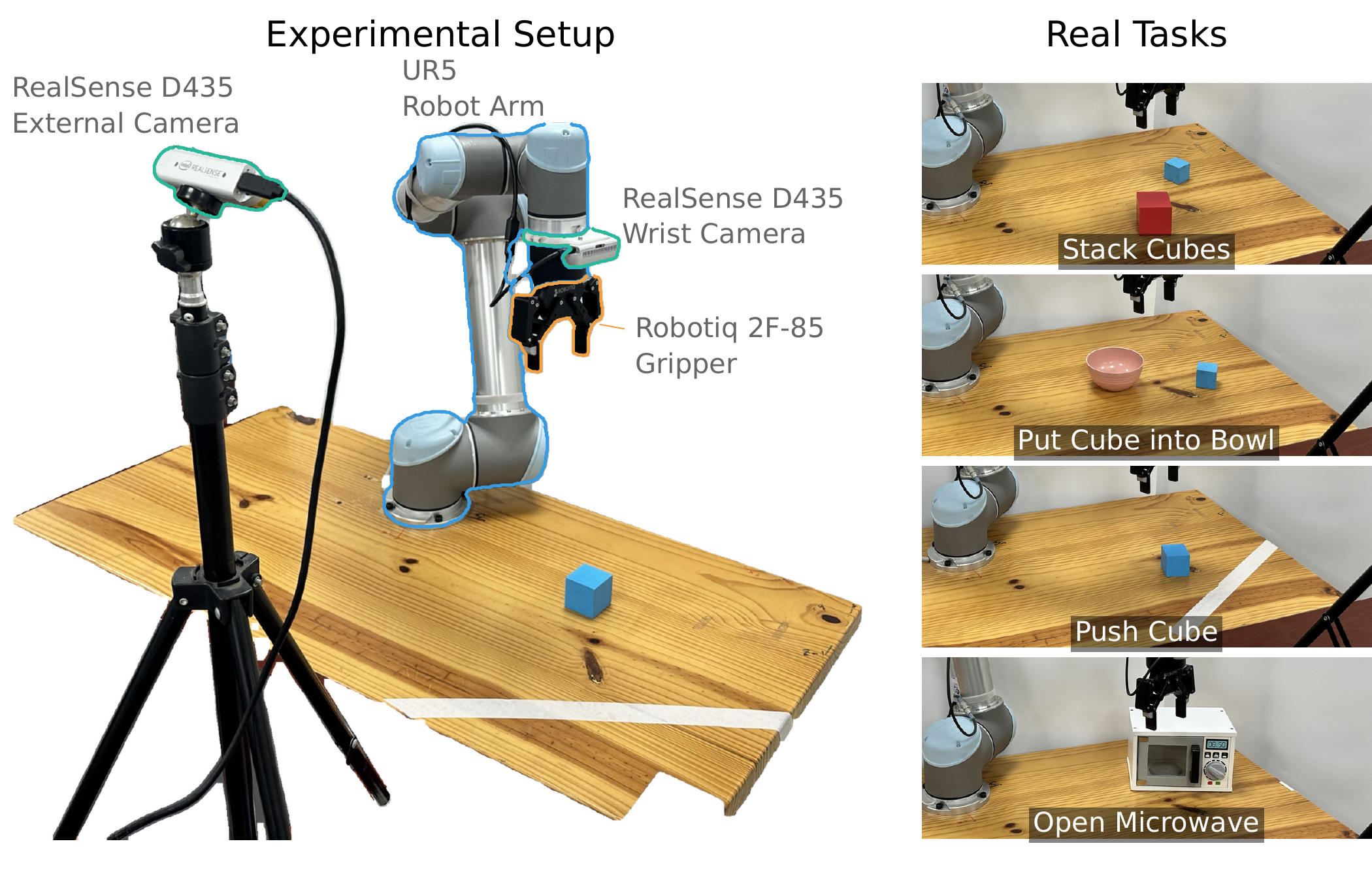}
  \caption{\textbf{Real-world experimental setup.} Left: the UR5 platform and camera configuration. Right: the four evaluation tasks.}
  \label{fig:real_world_setup}
\end{figure}

%% file: floats/real_policy.tex
\begin{table*}[t]
\centering\small
\setlength{\tabcolsep}{2pt}
\renewcommand{\arraystretch}{1.0}
\setlength{\aboverulesep}{1.5pt}
\setlength{\belowrulesep}{1.5pt}
\newcommand{\policyhead}[1]{\makebox[17pt][c]{#1}}
\caption{\textbf{Real-world policy success (\%).} S/T denote stage/full-task success. Averages cover the three rigid-object tasks for real-to-sim reconstruction and all four tasks elsewhere. Dashes indicate unavailable results; bold marks column-wise best results within each block.}
\label{tab:real_policy}
\begin{tabular*}{\textwidth}{@{\extracolsep{\fill}}ll*{10}{c}@{}}
\toprule
\textbf{Setting} & \textbf{Training data} &\multicolumn{2}{c}{\bfseries \shortstack{Stack\\Cubes}}&\multicolumn{2}{c}{\bfseries \shortstack{Put Cube\\into Bowl}}&\multicolumn{2}{c}{\bfseries \shortstack{Push\\Cube}}&\multicolumn{2}{c}{\bfseries \shortstack{Open\\Microwave}}&\multicolumn{2}{c}{\bfseries Average}\\
\cmidrule(lr){3-4}\cmidrule(lr){5-6}\cmidrule(lr){7-8}\cmidrule(lr){9-10}\cmidrule(lr){11-12}
&&\policyhead{S}&\policyhead{T}&\policyhead{S}&\policyhead{T}&\policyhead{S}&\policyhead{T}&\policyhead{S}&\policyhead{T}&\policyhead{S}&\policyhead{T}\\
\midrule
Real only & Real-50 & 30 & 15 & 30 & 20 & 60 & 45 & 10 & 10 & 32.5 & 22.5\\
\midrule
\multirow{3}{*}{\shortstack[l]{Real-to-sim\\reconstruction}} & Real-50 + RoboSnap~\cite{zhang2026robosnap}-R2S-50 & 20 & 15 & \textbf{45} & 15 & 45 & 25 & -- & -- & 36.7 & 18.3\\
 & Real-50 + AR2S~\cite{chen2026agentic}-R2S-50 & 25 & 15 & 35 & 20 & 50 & 35 & -- & -- & 36.7 & 23.3\\
 & Real-50 + OurR2S-50 & \textbf{35} & \textbf{20} & 40 & \textbf{40} & \textbf{60} & \textbf{55} & \textbf{45} & \textbf{30} & \textbf{45.0} & \textbf{38.3}\\
\midrule
\multirow{3}{*}{\shortstack[l]{Layout\\augmentation}} & Real-50 + OurR2S-50 + MimicGen~\cite{mandlekar2023mimicgen}-50 & 45 & 20 & 45 & 35 & 65 & 50 & 50 & 30 & 51.3 & 33.8\\
 & Real-50 + OurR2S-50 + CPGen~\cite{lin2025constraint}-50 & 50 & 35 & 45 & 40 & \textbf{85} & 50 & 60 & 35 & 60.0 & 40.0\\
 & Real-50 + OurR2S-50 + OurAug-50 & \textbf{55} & \textbf{50} & \textbf{60} & \textbf{55} & \textbf{85} & \textbf{60} & \textbf{65} & \textbf{50} & \textbf{66.3} & \textbf{53.8}\\
\midrule
\multirow{3}{*}{\shortstack[l]{Human-to-robot\\conversion}} & Real-50 + Ego2Robot~\cite{wang2026ego2robot}-50 & 40 & 20 & 40 & 35 & 60 & 50 & 30 & 25 & 42.5 & 32.5\\
 & Real-50 + OurH2R-50 & \textbf{65} & 45 & 55 & 50 & 75 & 60 & 45 & 30 & 60.0 & 46.3\\
 & Real-50 + OurH2R-50 + OurH2RAug-50 & \textbf{65} & \textbf{55} & \textbf{70} & \textbf{60} & \textbf{80} & \textbf{70} & \textbf{60} & \textbf{40} & \textbf{68.8} & \textbf{56.3}\\
\bottomrule
\end{tabular*}
\end{table*}

%% file: floats/ur5_reconstruction.tex
\begin{figure}[t]
\centering
\includegraphics[width=\columnwidth]{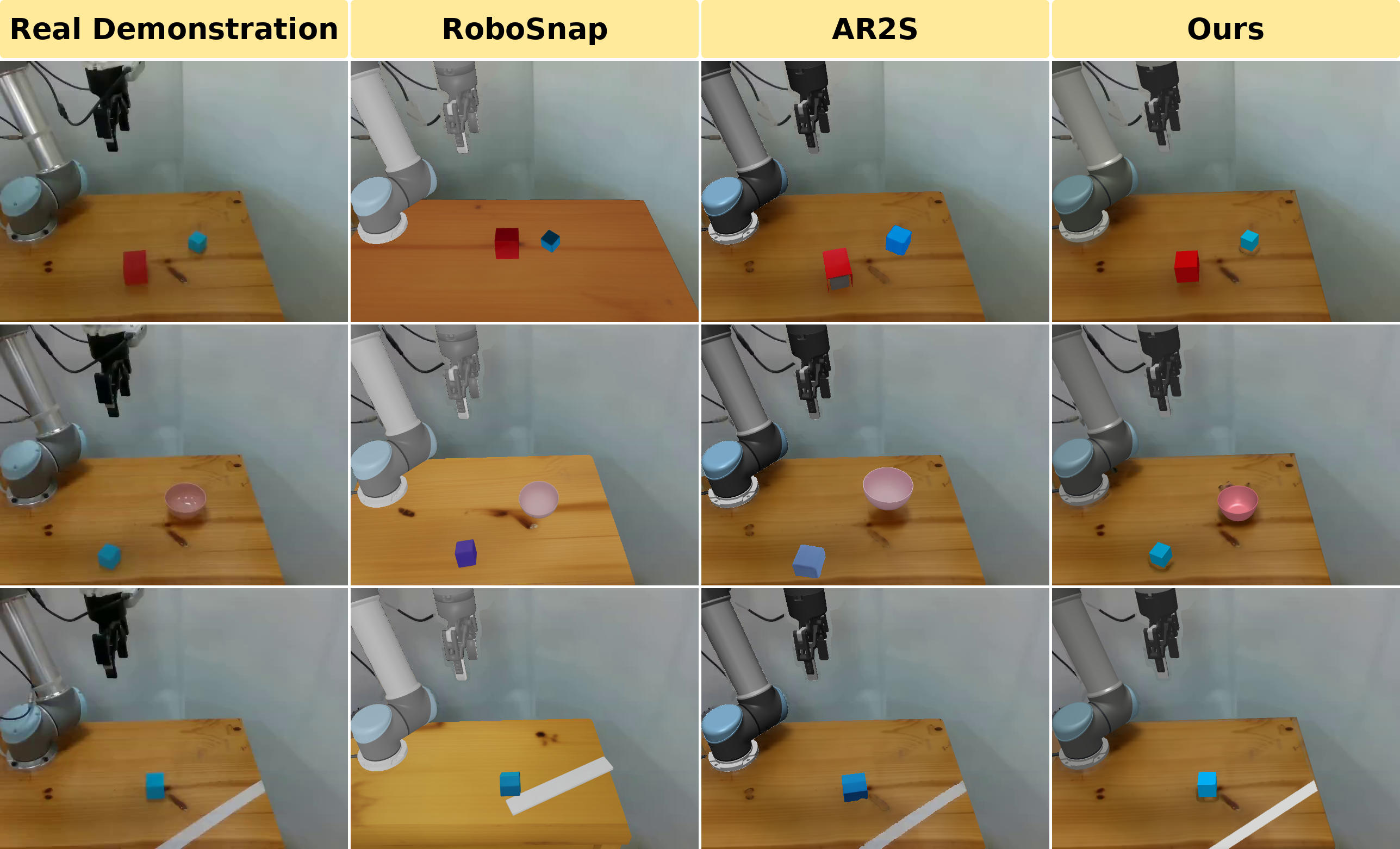}
\caption{\textbf{Real-to-sim reconstruction on UR5 tasks.} Each row pairs a frame from a real demonstration with the corresponding reconstructions for cube stacking, cube placement, or cube pushing.}
\label{fig:ur5_reconstruction}
\end{figure}

%% file: floats/bimanual_policy.tex
\begin{table}[!t]
\centering\footnotesize
\setlength{\tabcolsep}{1.5pt}
\renewcommand{\arraystretch}{1.0}
\setlength{\aboverulesep}{1.5pt}
\setlength{\belowrulesep}{1.5pt}
\caption{\textbf{Cross-simulator bimanual policy success (\%).} RoboTwin (RT), our reconstructions (S2S), DexMimicGen (DM), and our augmentations (OurAug) each supply 50 demos per task.}
\label{tab:bimanual_policy}
\footnotesize
\begin{tabular*}{\columnwidth}{@{\extracolsep{\fill}}lccccc@{}}
\toprule
\textbf{Training data}&\shortstack{\textbf{Pick Two}\\\textbf{Bottles}}&\shortstack{\textbf{Lift}\\\textbf{Pot}}&\shortstack{\textbf{Handover}\\\textbf{Microphone}}&\shortstack{\textbf{Put Bread}\\\textbf{into Pan}}&\textbf{Avg.}\\
\midrule
RT &30&39&49&28&36.5\\
RT+S2S &37&40&49&33&39.8\\
RT+S2S+DM~\cite{jiang2025dexmimicgen} &40&41&53&32&41.5\\
RT+S2S+OurAug &\textbf{45}&\textbf{49}&\textbf{54}&\textbf{37}&\textbf{46.3}\\
\bottomrule
\end{tabular*}
\par\vspace{\floatsep}
\begin{minipage}{\linewidth}
  \captionsetup{type=figure}
  \input{floats/ur5_object_adaptation}
\end{minipage}
\end{table}

%% file: floats/ur5_object_adaptation.tex
  \centering
\includegraphics[width=\linewidth]{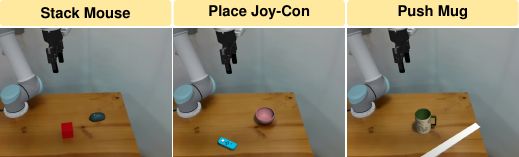}
\caption{\textbf{Target-object adaptation.} Simulated UR5 scenes after replacing the manipulated cube in each task.}
\label{fig:ur5_object_adaptation}

%% file: floats/policy_results_layout.tex
\begin{table}[!t]
  \begin{minipage}{\linewidth}
    \captionsetup{type=table}
    \input{floats/object_adaptation}
  \end{minipage}\par\vspace{\floatsep}
  \begin{minipage}{\linewidth}
    \captionsetup{type=table}
    \input{floats/robot_adaptation}
  \end{minipage}
\end{table}

%% file: floats/object_adaptation.tex
\centering\footnotesize
\setlength{\tabcolsep}{1.5pt}
\renewcommand{\arraystretch}{1.0}
\setlength{\aboverulesep}{1.5pt}
\setlength{\belowrulesep}{1.5pt}
\caption{\textbf{Real-world policy success on target objects (\%).} In task order, a mouse, a Joy-Con, and a mug replace the cube. S/T denote stage/full-task success. Best values in bold.}
\label{tab:object_adaptation}
\begin{tabular*}{\columnwidth}{@{\extracolsep{\fill}}lcccccccc@{}}
\toprule
\textbf{Training data}&\multicolumn{2}{c}{\textbf{Stack}}&\multicolumn{2}{c}{\textbf{Place}}&\multicolumn{2}{c}{\textbf{Push}}&\multicolumn{2}{c}{\textbf{Average}}\\
\cmidrule(lr){2-3}\cmidrule(lr){4-5}\cmidrule(lr){6-7}\cmidrule(lr){8-9}
&S&T&S&T&S&T&S&T\\
\midrule
Original-50&0&0&0&0&10&5&3.3&1.7\\
ObjAug-50&30&15&20&15&40&35&30.0&21.7\\
ObjAug-100&\textbf{50}&\textbf{30}&\textbf{35}&\textbf{25}&\textbf{55}&\textbf{45}&\textbf{46.7}&\textbf{33.3}\\
\bottomrule
\end{tabular*}

%% file: floats/robot_adaptation.tex
\centering\footnotesize
\setlength{\tabcolsep}{1.5pt}
\renewcommand{\arraystretch}{1.0}
\setlength{\aboverulesep}{1.5pt}
\setlength{\belowrulesep}{1.5pt}
\caption{\textbf{Real-world policy success on Piper-X (\%).} S/T denote stage/full-task success. Best values in bold.}
\label{tab:robot_adaptation}
\begin{tabular*}{\columnwidth}{@{\extracolsep{\fill}}lcccccccccc@{}}
\toprule
\textbf{Training data}&\multicolumn{2}{c}{\textbf{Stack}}&\multicolumn{2}{c}{\textbf{Place}}&\multicolumn{2}{c}{\textbf{Push}}&\multicolumn{2}{c}{\textbf{Open}}&\multicolumn{2}{c}{\textbf{Average}}\\
\cmidrule(lr){2-3}\cmidrule(lr){4-5}\cmidrule(lr){6-7}\cmidrule(lr){8-9}\cmidrule(lr){10-11}
&S&T&S&T&S&T&S&T&S&T\\
\midrule
RobotAug-50&45&20&\textbf{20}&10&\textbf{55}&20&\textbf{10}&0&32.5&12.5\\
RobotAug-100&\textbf{60}&\textbf{40}&\textbf{20}&\textbf{15}&50&\textbf{30}&\textbf{10}&\textbf{5}&\textbf{35.0}&\textbf{22.5}\\
\bottomrule
\end{tabular*}

%% file: floats/ur5_to_piper_x.tex
\begin{figure*}[t]
  \centering
  \includegraphics[width=\textwidth]{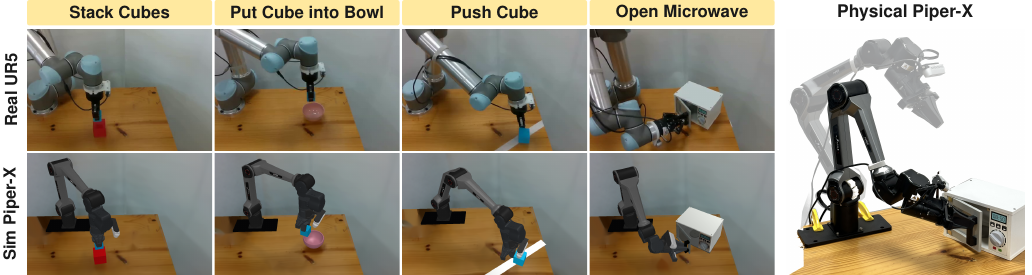}
  \caption{\textbf{Adapting UR5 demonstrations to Piper-X.} Each column pairs a real UR5 demonstration (top) with its simulated Piper-X counterpart (bottom). Policies trained solely on synthesized data are evaluated on the real Piper-X (right).}
  \label{fig:ur5_to_piper_x}
\end{figure*}

%% file: floats/trajectory_ablation.tex
\begin{table}[t]
\centering\footnotesize
\setlength{\tabcolsep}{1.5pt}
\renewcommand{\arraystretch}{1.0}
\setlength{\aboverulesep}{1.5pt}
\setlength{\belowrulesep}{1.5pt}
\caption{\textbf{Optimization ablations.} J-LDLJ and E-LDLJ measure joint and end-effector smoothness. Best values in bold.}
\label{tab:trajectory_ablation}
\footnotesize
\begin{tabular*}{\columnwidth}{@{\extracolsep{\fill}}lcccc@{}}
\toprule
\textbf{Configuration}&\mbox{\textbf{Success} (\%)\,$\uparrow$}&\mbox{\textbf{J-LDLJ}\,$\uparrow$}&\mbox{\textbf{E-LDLJ}\,$\uparrow$}&\mbox{\textbf{Time} (s)\,$\downarrow$}\\
\midrule
Initialization only&30&$-16.13$&$-16.03$&--\\
w/o scene updates&75&$-16.02$&$-15.93$&504\\
w/o targeted updates&90&$-16.00$&$-15.99$&388\\
\specialrule{0.3pt}{1.6pt}{1.6pt}
w/o checkpointing&\textbf{100}&$-15.95$&$-15.96$&315\\
w/o interpolation&\textbf{100}&$-16.29$&$-16.14$&280\\
\specialrule{0.3pt}{1.6pt}{1.6pt}
Ours&\textbf{100}&$\mathbf{-15.92}$&$\mathbf{-15.88}$&\textbf{223}\\
\bottomrule
\end{tabular*}
\end{table}

%% file: sec/5_conclusion.tex
\section{Conclusion}
\label{sec:conclusion}

RoboRecast generates synthetic data by preserving task-defining interaction requirements while adapting scenes and motions to target configurations.
It extracts these requirements from source demonstrations and uses simulation feedback to diagnose failures and identify which scene parameters or robot motions to revise.
Adding our synthetic data to target demonstrations improved policy success on real single-arm and simulated bimanual tasks. Without target demonstrations, it enabled training for new objects or robots.
These results support interaction requirements as a useful abstraction for preserving manipulation intent while allowing flexible adaptation across configurations.
We discuss limitations and future work in detail in the appendix.

%% file: sec/6_llm_usage_statement.tex
\section*{LLM Usage Statement}
We used Codex with GPT-5 and GPT-6 to assist with manuscript drafting and revision, identifying related work, code implementation, and running and debugging experiments under author direction and supervision.
The RoboRecast, AR2S, and RoboSnap data generation pipelines each require calls to a vision-language model or an agent during generation.
In our comparison, we used GPT-6-Astra as the underlying model for these calls in all three pipelines.
We also evaluated the RoboRecast agent with GPT-5.6-Sol, GPT-6.1-Sol, and GPT-6-Astra, and results of this model comparison are in the appendix.
The authors developed the research ideas, method, tool interfaces, and experimental protocols, reviewed and tested the AI-assisted code, and manually verified AI-assisted content and experimental outputs.
The authors take responsibility for the final content and results of this work.

%% file: sec/appendix/contents.tex
\begingroup
\small
\setcounter{tocdepth}{2}
\noindent\begin{minipage}{\columnwidth}
\textbf{Contents}\par\vspace{4pt}
\makeatletter
\renewcommand{\l@section}[2]{\addvspace{3pt}{\bfseries\@dottedtocline{1}{0em}{1.6em}{#1}{#2}}}
\renewcommand{\l@subsection}{\@dottedtocline{2}{1.0em}{2.3em}}
\@starttoc{toc}
\makeatother
\end{minipage}\par
\endgroup

%% file: sec/appendix/a_implementation.tex
\section{Implementation Details}
\label{sec:app_implementation}

\subsection{Skills and Tools}
\label{sec:app_skills_tools}
Table~\ref{tab:app_skills_tools} lists the skills and tools we provide to RoboRecast, our agentic optimization framework for demonstration augmentation.
Skills guide the agent in interpreting demonstrations, constructing candidate scenes and motions, and deciding how to revise them when a check or rollout fails.
The agent invokes programmatic tools for modeling, motion compilation, validity checks, and simulation, and interprets their outputs to determine what to change.
\input{floats/appendix/skills_tools}

\subsection{Method Details}
\noindent\textbf{Parallel modeling of scene units.}
The agent uses the source scene graph to group parts that must fit together into modeling units, while assigning independently movable objects to separate units.
For example, the countertop, inset sink, and faucet form a single modeling unit so that the sink fits the countertop opening and the faucet is positioned over the basin.
Modeling these fixed assemblies together preserves their attachment relationships, avoiding unintended tipping or displacement during settling that can arise when attached parts are treated as independent objects.

To reduce scene modeling time, subagents process different units in parallel, following AgentSTAR's workflow~\cite{mazur2026agentstar} to alternate between reconstructing geometry and appearance and refining pose estimates.
Each subagent also specifies its unit's collision geometry, physical properties, and joints where applicable, as described in Sec.~3.2.
The agent then assembles the units according to their poses and the scene graph, and checks that the assembled scene preserves the observed support and contact relationships.

\noindent\textbf{Compiling semantic waypoints.}
The three motion modes defined in Sec.~3.1---\textit{transit}, \textit{dwell}, and \textit{sustained contact}---share an interpolation scheme but differ in how their waypoints are placed.
Transit waypoints connect poses in free space, dwell waypoints hold the end effector at a fixed pose for a specified duration, and sustained-contact waypoints trace a path defined by the interaction geometry, such as the arc of a door about its hinge.
For the segment from waypoint $j-1$ to waypoint $j$, let $u=(t-t_{j-1})/\Delta t_j$ denote normalized time.
After transforming the waypoint positions into the world frame, denoted by $\bar p_j$, the compiler interpolates between successive positions using a quintic time law~\cite{lynch2017modern},
\begin{equation}
\begin{aligned}
s(u) &= 10u^3-15u^4+6u^5,\\
p(t) &= \bar p_{j-1}+s(u)\bigl(\bar p_j-\bar p_{j-1}\bigr).
\end{aligned}
\end{equation}
For fixed endpoint positions and segment duration, this time law minimizes integrated squared translational jerk subject to zero endpoint velocity and acceleration~\cite{flash1985coordination}.
The resulting position trajectory has continuous velocity and acceleration across waypoint boundaries.

The compiler also uses $s(u)$ to interpolate between the gripper widths specified by the agent at consecutive waypoints, producing smooth opening and closing commands with zero endpoint velocity and acceleration.
To obtain similarly smooth orientation changes, it applies spherical linear interpolation (SLERP)~\cite{shoemake1985animating} to world-frame waypoint quaternions $\bar q_j$ using the same time law,
\begin{equation}
q(t)=\operatorname{SLERP}\!\left(\bar q_{j-1},\bar q_j,s(u)\right),
\label{eq:app_orientation_interpolation}
\end{equation}
giving zero angular velocity and acceleration at both ends of each segment.
The compiler solves inverse kinematics for each sampled pose in sequence, using the preceding joint solution to initialize the next solve.

\noindent\textbf{Scene and motion validity checks.}
Before rollout, the agent invokes our programmatic tools to check scene and motion validity, then analyzes their outputs.
For scene validity, the agent compares rendered views with source observations and the target configuration to check object appearance, placement, and asset replacement.
Collision geometry is overlaid on visual meshes in three orthographic views to reveal missing surfaces or obstructed interaction regions, such as handles and container openings.
A settling simulation exposes unsupported objects, unintended tipping, unstable placement, and interpenetration.
For articulated objects, the agent invokes tools to check joint axes and limits and test for blocked joint travel.

For motion validity, the tools compare forward-kinematics poses of the computed joint configurations with the requested end-effector poses to identify unreachable or inaccurately realized targets.
They also check joint positions and velocities between successive samples against the robot's limits.
All checks run on the initial candidate and are repeated as needed after edits.
For example, after editing a hinged object's collision geometry, the agent inspects the updated overlay and invokes the joint-motion check to verify that the geometry does not obstruct rotation.

\noindent\textbf{Bimanual manipulation.}
For each phase, the agent specifies when each gripper should establish, maintain, or release contact.
When both arms carry an object, it constructs their waypoints from the same sequence of desired object poses and their respective grasp locations.
During rollout, the agent verifies contact timing, and collision checks flag unintended contacts within each arm, between the arms, and with the scene.
For example, if an edit delays the receiving arm during a handover, the agent extends the delivering arm's hold and verifies that it releases the object only after the receiving gripper has secured it.

\input{floats/appendix/human_to_robot}
\noindent\textbf{Human-to-robot conversion.}
Hand poses provide the agent with reference information about how the demonstrator interacts with objects, guiding its construction of semantic waypoints for the target robot.
These poses can come from dataset annotations or off-the-shelf estimators such as WiLoR~\cite{potamias2025wilor}, depending on the available input.
Together with the video, the thumb and index fingertip positions and their relative motion help the agent infer contact regions, grasping directions, and the timing of grasp and release (Fig.~\ref{fig:app_human_to_robot}).
The agent uses these cues alongside the interaction requirements, target object geometry, and robot reachability to choose grasp poses, approach paths, and gripper commands when initializing the waypoints.
Our current pipeline supports tasks realizable with parallel-jaw grippers, excluding demonstrations whose essential interactions require dexterous manipulation.
One of the authors performed all human demonstrations collected for this study and consented to their use in the experiments and to publication of the images in Fig.~\ref{fig:app_human_to_robot}.

\subsection{Training Data Generation and Export}
Once a MuJoCo~\cite{todorov2012mujoco} rollout is accepted, we convert it into an observation--action sequence tailored to the downstream policy.
For the Diffusion Policies~\cite{chi2025diffusion} used in our experiments, we sample external-camera images, actual robot joint positions, and gripper widths from the recorded rollout at the policy's control frequency.
Each observation is paired with the subsequent control command, which specifies the target joint positions and gripper width.
These observation and action conventions match those of the real demonstrations, allowing synthetic and real episodes to be used by the same policy training pipeline.

To generate the visual observations, we restore the saved simulation states and render each frame from the calibrated external camera.
The background can be represented using Gaussian splats generated by InfiniSplat~\cite{wang2026infinisplat} or incorporated through 2D compositing.
Geometry-based visibility masks preserve the rendered robot and task objects when integrating the background.
The resulting images are resized to the policy's input resolution and stored alongside the corresponding robot states and actions.

We trim inactive intervals at the beginning and end of each episode while retaining pauses within the task, applying the same temporal crop to images, states, and actions.
Episodes generated through reconstruction, layout augmentation, or human-to-robot conversion share the same data format, with joint ordering and gripper-width mappings adapted to the target robot.
Before export, we check that timestamps align, image and state counts match, and action labels agree with logged control targets.

%% file: floats/appendix/skills_tools.tex
\begin{table*}[t]
\centering\small
\renewcommand{\arraystretch}{1.08}
\caption{\textbf{Skills and tools in RoboRecast.}}
\label{tab:app_skills_tools}
\begin{tabular}{@{}p{.17\textwidth}@{\hspace{1em}}p{.37\textwidth}@{\hspace{1em}}p{\dimexpr.46\textwidth-2em\relax}@{}}
\toprule
\textbf{Stage} & \textbf{Skills} & \textbf{Tools} \tabularnewline
\midrule
\raggedright Task Intent Analysis &
\raggedright Identify object roles and required interactions;
map contacts, state changes, and their ordering to the target configuration &
\raggedright Keyframe and clip extraction;
observation visualization;
SAM~3 segmentation~\cite{carion2026sam};
LingBot-Depth refinement~\cite{tan2026masked} \tabularnewline
\addlinespace
\raggedright Scene initialization &
\raggedright Group attached parts into modeling units in the scene graph;
assemble units for the target configuration &
\raggedright AgentSTAR Blender modeling, rendering, and pose estimation~\cite{mazur2026agentstar};
convex decomposition~\cite{shi2026cuacd};
collision primitives;
physical and joint parameter setup;
scene assembly \tabularnewline
\addlinespace
\raggedright Motion initialization and compilation &
\raggedright Choose contact regions and approach directions;
specify waypoint motion modes and timing &
\raggedright Frame transformations;
quintic position and gripper-width interpolation;
spherical orientation interpolation;
sequential inverse kinematics \tabularnewline
\addlinespace
\raggedright Validity checking &
\raggedright Interpret scene validity checks;
interpret motion feasibility checks &
\raggedright Three-view collision overlays;
settling and penetration tests;
joint travel checks;
pose error and joint-limit checks \tabularnewline
\addlinespace
\raggedright Diagnosis-driven optimization &
\raggedright Locate the earliest unmet requirement or physical violation;
select a targeted scene or waypoint update;
verify the repair &
\raggedright MuJoCo rollouts~\cite{todorov2012mujoco};
rendered keyframes;
contact and state logs;
scene and waypoint editing;
checkpoint saving and restoration \tabularnewline
\bottomrule
\end{tabular}
\end{table*}

%% file: floats/appendix/human_to_robot.tex
\begin{figure}[t]
\centering
\begingroup
\definecolor{h2rheader}{HTML}{FFE99F}
\setlength{\fboxsep}{0pt}
\newcommand{\hTwoRheading}[1]{\colorbox{h2rheader}{\parbox[c][12.5pt][c]{0.33126\columnwidth}{\centering\sffamily\bfseries\fontsize{7}{8}\selectfont #1}}}
\noindent\hTwoRheading{Stack Cubes}\hfill\hTwoRheading{Place cube}\hfill\hTwoRheading{Open Microwave}\par
\vspace{0.8pt}
\noindent\includegraphics[width=\columnwidth]{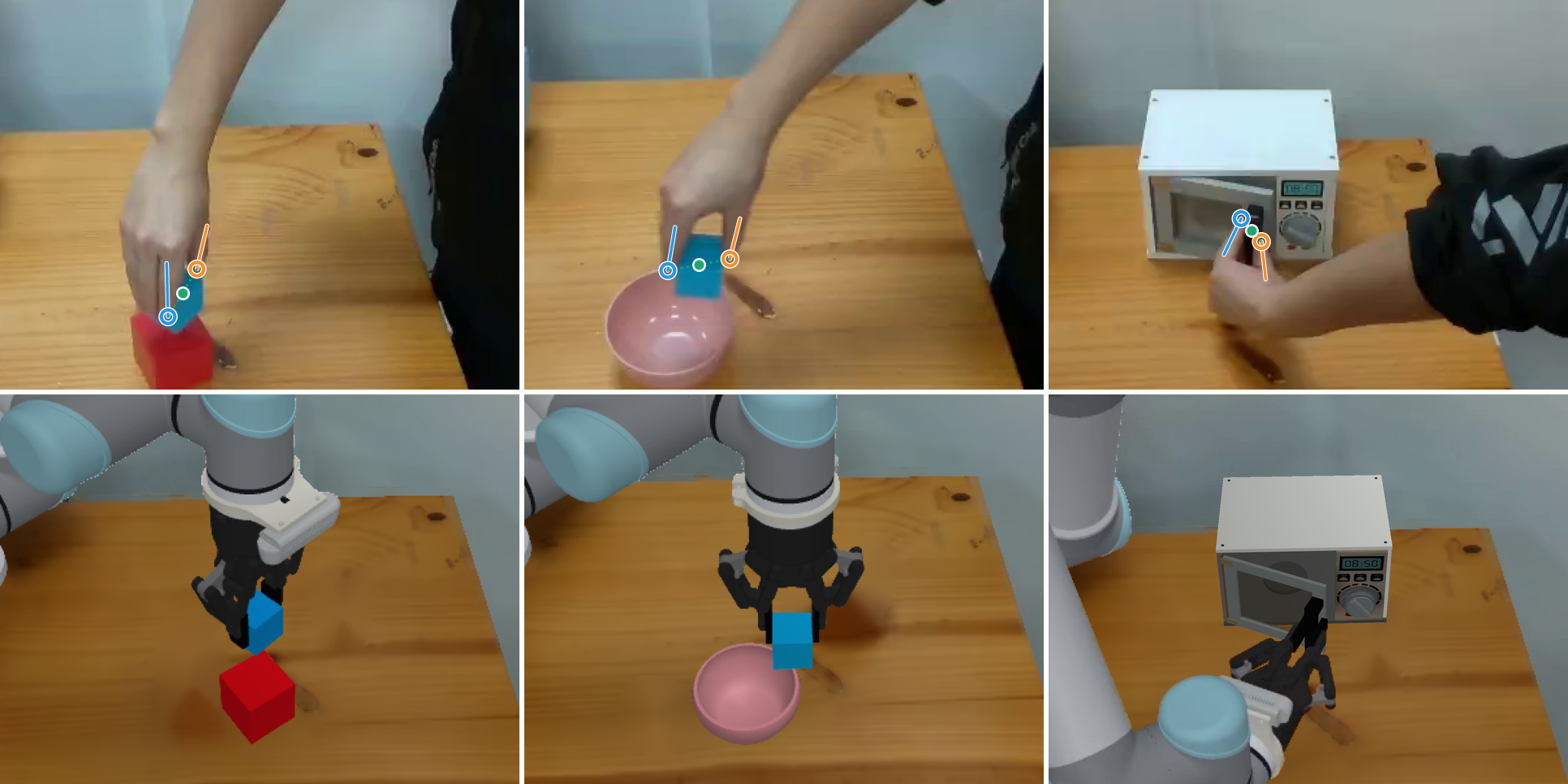}
\endgroup
\caption{\textbf{Human-to-robot conversion examples.} Top: human demonstrations with fingertip references and their midpoints. Bottom: corresponding robot actions in simulation.}
\label{fig:app_human_to_robot}
\end{figure}

%% file: sec/appendix/b_evaluation_protocols_and_additional_results.tex
\section{Evaluation Details and Additional Results}
\label{sec:app_evaluation}

\subsection{Evaluation Protocols and Metrics}
\label{sec:app_protocols}
\noindent\textbf{Policy training.} We train Diffusion Policy~\cite{chi2025diffusion} using external-camera images and robot proprioception as observations.
A ResNet-18~\cite{he2016deep} encoder trained from scratch encodes the images, and a conditional 1D U-Net predicts joint-position and gripper-width target sequences.
Across all policy evaluations, we train each policy for 600 epochs using the same architecture, random seed, and optimizer settings within each comparison.
Table~\ref{tab:app_policy_training} summarizes the training configuration.
\input{floats/appendix/policy_training}

\noindent\textbf{Policy evaluation.} We evaluate each policy over 20 trials per task on the physical robots and 100 trials per task in RoboTwin~\cite{chen2025robotwin}, using the same predefined initial configurations across methods.
On UR5 and Piper-X, a trial fails if the policy does not complete the task within 60 seconds or triggers the robot's protective stop.
Full-task success requires completing the task goal, whereas stage success records whether the policy achieves a designated intermediate interaction.
For stacking, placement, and microwave opening, stage success requires grasping the manipulated object or door handle.
For pushing, it requires moving the object partway toward the goal.
We report the arithmetic mean of per-task success rates, specifying the evaluated task subset whenever methods do not cover all tasks.

\noindent\textbf{Demonstration generation.} We evaluate generation success by whether the synthesized rollout completes the required task interactions in simulation.
The agent determines when to stop refinement, and we manually review the final rollout to assess task completion.
Generation success is the fraction of evaluated inputs that yield a successful rollout within the prescribed budget, including unsuccessful attempts in the denominator.

\noindent\textbf{Trajectory smoothness.} We quantify motion smoothness using log dimensionless jerk (LDLJ) and spectral arc length (SPARC), with higher values indicating smoother motion for both metrics~\cite{balasubramanian2015analysis}.
For a scalar velocity signal $v(t)$ over duration $T$, LDLJ measures changes in acceleration after normalization by duration and peak velocity:
\begin{equation}
\operatorname{LDLJ}(v)=-\ln\!\left[\frac{T^3}{v_{\mathrm{peak}}^2}\int_0^T\left(\frac{d^2v(t)}{dt^2}\right)^2dt\right].
\label{eq:app_ldlj}
\end{equation}
Here, $v_{\mathrm{peak}}=\max_t|v(t)|$, making the measure invariant to uniform temporal and amplitude scaling.
SPARC instead measures the negative arc length of the normalized magnitude spectrum of the speed signal:
\begin{equation}
\operatorname{SPARC}=-\int_0^{f_c}\sqrt{\frac{1}{f_c^2}+\left(\frac{d\widehat V(f)}{df}\right)^2}\,df.
\label{eq:app_sparc}
\end{equation}
Here, $\widehat V(f)$ is the Fourier magnitude spectrum of $|v(t)|$, normalized by its maximum, and $f_c$ is an adaptive cutoff selected using a spectral amplitude threshold and a maximum frequency.

We compute these metrics from the executed robot trajectories rather than the planned waypoints.
J-LDLJ is computed from each active arm joint's signed angular velocity and averaged across joints, excluding the gripper.
E-LDLJ is computed from the end-effector translational speed, $v(t)=\|\dot{\mathbf p}(t)\|_2$.
For SPARC, we use the absolute angular velocity for each joint and the translational speed for the end effector.
We evaluate trajectories at 20 Hz without filtering and include only arm joints with an angular range of at least $1^\circ$ and a peak absolute angular velocity of at least $0.01\,\mathrm{rad/s}$.
SPARC uses a maximum cutoff frequency of 10 Hz, a normalized spectral amplitude threshold of 0.05, and a zero-padding level of 4.
For the optimization ablations in the main paper, we average scores over all final rollouts, including both successful and failed executions.

\noindent\textbf{Computational cost.} End-to-end generation time covers initialization and refinement through final rollout assessment, excluding subsequent training-data rendering and export.
Optimization time starts after scene and waypoint initialization.
We report mean wall-clock time over all evaluated inputs, including model inference, tool execution, and unsuccessful refinement attempts.

Since experiments run through a Codex subscription, we report reference-price-equivalent model usage cost rather than actual charges.
Token counts include the task's main agent and internal subagents.
Using the evaluated model's rates from OpenAI's official API pricing, we estimate cost in USD as
\begin{equation}
C_{\mathrm{ref}}=10^{-6}\bigl(r_uN_u+r_cN_c+r_oN_o\bigr),
\label{eq:app_reference_cost}
\end{equation}
where $r_k$ and $N_k$ denote the rate in USD per million tokens and the token count for uncached input ($u$), cached input ($c$), and output ($o$).
We compute $N_u=N_{\mathrm{input}}-N_c$; reasoning tokens are already included in $N_o$ and are not counted separately.
This estimate excludes GPU, reconstruction-model, simulation, and storage costs.

\subsection{Additional Policy Evaluation Results}
\label{sec:app_policy}
\input{floats/appendix/ur5_additional_policy}

\noindent\textbf{Training on synthetic demonstrations alone.} We use real UR5 demonstrations as reconstruction inputs and evaluate policies trained solely on the resulting synthetic data (Table~\ref{tab:app_ur5_additional_policy}, top).
Across four tasks, training on 50 reconstructed trajectories per task yields 13.8\% mean full-task success, compared with 22.5\% for 50 real demonstrations.
Adding 50 layout-augmented trajectories per task to the reconstructed set raises mean success to 30.0\%.
These results suggest that trajectories rendered from reconstructed scenes with appropriate background integration can provide useful supervision for real-world manipulation.

\noindent\textbf{Augmentation from different reconstructions.} To examine how reconstruction quality affects subsequent augmentation, we apply RoboRecast's layout augmentation to reconstructions produced by RoboSnap~\cite{zhang2026robosnap}, AR2S~\cite{chen2026agentic}, and RoboRecast.
Each policy is trained on 50 real demonstrations, 50 reconstructed trajectories, and 50 layout-augmented trajectories per task.
Applying RoboRecast to RoboSnap and AR2S reconstructions raises mean full-task success on the three rigid-object tasks from 18.3\% to 28.3\% and from 23.3\% to 30.0\%, respectively (Table~\ref{tab:app_ur5_additional_policy}).
These gains suggest that RoboRecast's diagnosis-driven optimization can partially compensate for inaccuracies in the reconstructed scenes by refining both scene parameters and robot motion, yielding more effective training data.

\input{floats/appendix/bimanual_object_adaptation}
\input{floats/appendix/bimanual_object_examples}
\noindent\textbf{Bimanual adaptation to new objects.} We next evaluate bimanual policy learning on new objects for which no demonstrations are available.
Across the four RoboTwin tasks, we replace the source bottles, pots, microphones, and bread with new assets and use RoboRecast to adapt the source demonstrations to these target objects (Fig.~\ref{fig:app_bimanual_object_examples}).
Policies trained on 50 synthesized trajectories per task achieve 36.25\% mean success on the target assets, compared with 18.25\% for policies trained on 50 original RoboTwin demonstrations (Table~\ref{tab:app_bimanual_object_adaptation}).
Increasing the synthesized training set to 100 trajectories per task improves performance on all four tasks, raising mean success to 47.75\%.
These results show that adapting source interactions to new objects provides effective training data for bimanual tasks without collecting target demonstrations.

\input{floats/appendix/visual_alignment_metrics}
\input{floats/appendix/single_arm_augmentation}
\subsection{Reconstruction Quality}
\label{sec:app_quality}
We compare the initial scene reconstruction quality of RoboRecast, RoboSnap~\cite{zhang2026robosnap}, and AR2S~\cite{chen2026agentic} on 20 DROID~\cite{khazatsky2024droid} demonstrations.
All methods reconstruct the foreground and share the same InfiniSplat~\cite{wang2026infinisplat} background, with metrics computed over the full images.
We measure pixel fidelity with PSNR, structural similarity with SSIM~\cite{wang2004ssim}, perceptual similarity with LPIPS~\cite{zhang2018perceptual}, feature correspondence with SIFT-MR~\cite{lowe2004sift}, and color distribution differences with RGB-EMD~\cite{rubner2000emd}.
SIFT-MR normalizes matches passing a 0.75 ratio test by the smaller keypoint count; RGB-EMD averages one-dimensional Wasserstein distances between 64-bin histograms of the three color channels.
RoboRecast achieves the best average scores across all five metrics (Table~\ref{tab:app_visual_alignment}).
The qualitative comparisons in Fig.~\ref{fig:visual_alignment} show closer correspondence in object appearance and spatial arrangement, complementing the policy evaluations with direct evidence of visual reconstruction quality.

\subsection{Demonstration Augmentation}
\label{sec:app_augmentation}
\input{floats/appendix/object_augmentation}
\input{floats/appendix/bimanual_layout_augmentation}
\noindent\textbf{Evaluation settings.} We evaluate whether RoboRecast generates successful, smooth demonstrations under changes in object layout, object geometry, and robot embodiment.
Source demonstrations come from the simulation datasets of MimicGen~\cite{mandlekar2023mimicgen} for single-arm experiments and DexMimicGen~\cite{jiang2025dexmimicgen} for bimanual experiments.
We compare against MimicGen and SkillGen~\cite{garrett2024skillmimicgen} for single-arm layout, object, and robot augmentation, and DexMimicGen for bimanual layout augmentation.
Figure~\ref{fig:app_augmentation_examples} illustrates source demonstrations and their RoboRecast-generated counterparts in these four settings.

\input{floats/visual_alignment}
\input{floats/appendix/augmentation_examples}

We follow MimicGen's layout (D0--D2), object (O1/O2), and robot-transfer settings, using Sawyer, IIWA, and UR5e as target robots.
For bimanual augmentation, we extend DexMimicGen's default layout distribution (P0) to broader distributions (P1/P2).
Within each comparison, all methods use the same source demonstration pool and target test configurations.
We run 20 trials per method and setting for layout and robot augmentation, and 30, 30, and 60 trials per method for object augmentation under D0, O1, and O2, respectively.
We report generation success, determined by whether the environment's success condition is reached during the final delivered rollout, and joint motion smoothness measured by J-LDLJ and J-SPARC (Sec.~\ref{sec:app_protocols}).

\noindent\textbf{Single-arm layout augmentation.} We evaluate single-arm layout augmentation on Square, Assembly, and Coffee using the D0, D1, and D2 initial-state distributions.
D0 follows the source demonstration distribution, while D1 and D2 introduce progressively larger variations in initial object positions.
RoboRecast achieves 100\% generation success across all three tasks and distributions, substantially outperforming MimicGen and SkillGen (Table~\ref{tab:app_single_arm_augmentation}, top).
The generated trajectories also attain higher J-LDLJ and J-SPARC scores on all three tasks, indicating smoother joint motion.

\noindent\textbf{Object augmentation.} We next evaluate adaptation to different object geometries on Mug Cleanup, where the robot places a mug into a drawer and closes it.
We compare the original mug (D0), a fixed replacement mug (O1), and mugs sampled from a pool of 12 assets (O2), all using the D0 initial-pose distribution.
RoboRecast achieves 100\% generation success in all three settings and higher J-LDLJ and J-SPARC scores than MimicGen and SkillGen (Table~\ref{tab:app_object_augmentation}).

\noindent\textbf{Robot augmentation.} We evaluate robot augmentation on Coffee by adapting demonstrations collected with Panda to Sawyer, IIWA, and UR5e.
Each target robot is evaluated under D0, D1, and D2, testing adaptation to both robot kinematics and initial object positions.
RoboRecast achieves 100\% generation success across all target robots and distributions, with higher J-LDLJ and J-SPARC scores than MimicGen and SkillGen (Table~\ref{tab:app_single_arm_augmentation}, bottom).

\noindent\textbf{Bimanual layout augmentation.} We further evaluate bimanual layout augmentation on Threading, Assembly, and Transport, comparing RoboRecast with DexMimicGen.
Starting from the default distribution P0, we construct P1 and P2 with progressively larger variations in object positions and orientations.
RoboRecast achieves 100\% generation success across all three tasks and distributions, outperforming DexMimicGen in both generation success and joint motion smoothness (Table~\ref{tab:app_bimanual_layout_augmentation}).

Together, these experiments support formulating augmentation around the required interactions, allowing grasp contacts and motion paths to adapt to new layouts, object geometries, and robot embodiments.
By revising these choices through simulation feedback, RoboRecast generates demonstrations with higher execution success and smoother motion than the evaluated rule-based baselines.

\subsection{Generation Cost}
\label{sec:app_ablations}
\input{floats/appendix/generation_cost}
We compare GPT-5.6-Sol, GPT-6.1-Sol, and GPT-6-Astra as the agent backbone to assess how model choice affects generation time and computational cost, using high reasoning effort for all three.
Each model processes the same five DROID~\cite{khazatsky2024droid} demonstrations: four articulated-object tasks involving a microwave, an oven, a drawer, and a rotating faucet, and one rigid-object task involving placing a tomato into a bowl.
Each run continues until successful completion, without a fixed time or refinement-round limit.
We report mean scene initialization time (Init.), joint optimization time (Opt.), total generation time, and equivalent API cost (USD) over all five cases.

All three models complete all five tasks, but GPT-5.6-Sol takes substantially longer in both initialization and joint optimization, with a mean total time of 83.53 minutes (Table~\ref{tab:app_generation_cost}).
GPT-6-Astra is slightly faster overall than GPT-6.1-Sol (26.84 versus 29.01 minutes), but GPT-6.1-Sol incurs substantially lower equivalent API cost (\$3.10 versus \$21.52 per case).
On these five cases, GPT-6.1-Sol offers the lowest model usage cost while achieving a generation time comparable to GPT-6-Astra.

%% file: floats/appendix/policy_training.tex
\begin{table}[t]
\centering\footnotesize
\setlength{\tabcolsep}{3pt}
\renewcommand{\arraystretch}{1.0}
\caption{\textbf{Policy training configuration.}}
\label{tab:app_policy_training}
\begin{tabular*}{\columnwidth}{@{\extracolsep{\fill}}ll@{}}
\toprule
\textbf{Parameter} & \textbf{Value}\\
\midrule
Visual encoder & ResNet-18, from scratch\\
Denoising network & Conditional 1D U-Net\\
U-Net channels & 512, 1024, 2048\\
Image resolution (W$\times$H) & $320\times240$\\
Proprioception / action dim. & 7 / 7\\
Nominal control frequency & 10 Hz\\
Obs. / pred. / action horizon & 2 / 16 / 8\\
Training diffusion steps & 100\\
Inference sampler & DDIM, 16 steps\\
\midrule
Optimizer & AdamW\\
Learning rate / weight decay & $10^{-4}$ / $10^{-6}$\\
Learning-rate schedule & Cosine\\
Warmup & 500 optimizer steps\\
Batch size & 64\\
Training epochs & 600\\
Random seed & 42\\
EMA & Enabled\\
Training hardware & One NVIDIA H100 80GB\\
\bottomrule
\end{tabular*}
\end{table}

%% file: floats/appendix/ur5_additional_policy.tex
\begin{table*}[t]
\centering\small
\setlength{\tabcolsep}{2pt}
\renewcommand{\arraystretch}{1.0}
\setlength{\aboverulesep}{1.5pt}
\setlength{\belowrulesep}{1.5pt}
\newcommand{\appurhead}[1]{\makebox[17pt][c]{#1}}
\caption{\textbf{Additional real-world UR5 policy results (\%).} S/T: stage/full-task success. Means cover four tasks in the first block and three rigid-object tasks elsewhere. Dashes denote unavailable results; best values per column within each block are bold.}
\label{tab:app_ur5_additional_policy}
\begin{tabular*}{\textwidth}{@{\extracolsep{\fill}}l*{10}{c}@{}}
\toprule
\textbf{Training data}&\multicolumn{2}{c}{\bfseries\shortstack{Stack\\Cubes}}&\multicolumn{2}{c}{\bfseries\shortstack{Put Cube\\into Bowl}}&\multicolumn{2}{c}{\bfseries\shortstack{Push\\Cube}}&\multicolumn{2}{c}{\bfseries\shortstack{Open\\Microwave}}&\multicolumn{2}{c}{\bfseries Average}\\
\cmidrule(lr){2-3}\cmidrule(lr){4-5}\cmidrule(lr){6-7}\cmidrule(lr){8-9}\cmidrule(lr){10-11}
&\appurhead{S}&\appurhead{T}&\appurhead{S}&\appurhead{T}&\appurhead{S}&\appurhead{T}&\appurhead{S}&\appurhead{T}&\appurhead{S}&\appurhead{T}\\
\midrule
OurR2S-50 &20&15&30&10&40&25&10&5&25.0&13.8\\
Real-50 &30&15&30&\textbf{20}&60&45&10&10&32.5&22.5\\
OurR2S-50 + OurAug-50 &\textbf{45}&\textbf{25}&\textbf{45}&\textbf{20}&\textbf{65}&\textbf{50}&\textbf{35}&\textbf{25}&\textbf{47.5}&\textbf{30.0}\\
\midrule
Real-50 + RoboSnap~\cite{zhang2026robosnap}-R2S-50 &20&\textbf{15}&\textbf{45}&15&45&25&--&--&36.7&18.3\\
Real-50 + RoboSnap~\cite{zhang2026robosnap}-R2S-50 + OurAug-50 &\textbf{40}&\textbf{15}&35&\textbf{30}&\textbf{60}&\textbf{40}&--&--&\textbf{45.0}&\textbf{28.3}\\
\midrule
Real-50 + AR2S~\cite{chen2026agentic}-R2S-50 &25&15&\textbf{35}&\textbf{20}&50&35&--&--&36.7&23.3\\
Real-50 + AR2S~\cite{chen2026agentic}-R2S-50 + OurAug-50 &\textbf{45}&\textbf{30}&25&\textbf{20}&\textbf{60}&\textbf{40}&--&--&\textbf{43.3}&\textbf{30.0}\\
\midrule
Real-50 + OurR2S-50 &35&20&40&40&60&55&45&30&45.0&38.3\\
Real-50 + OurR2S-50 + OurAug-50 &\textbf{55}&\textbf{50}&\textbf{60}&\textbf{55}&\textbf{85}&\textbf{60}&\textbf{65}&\textbf{50}&\textbf{66.7}&\textbf{55.0}\\
\bottomrule
\end{tabular*}
\end{table*}

%% file: floats/appendix/bimanual_object_adaptation.tex
\begin{table}[t]
\centering\footnotesize
\setlength{\tabcolsep}{1.5pt}
\renewcommand{\arraystretch}{1.0}
\setlength{\aboverulesep}{1.5pt}
\setlength{\belowrulesep}{1.5pt}
\caption{\textbf{Cross-simulator bimanual policy success on new objects (\%).} Replace denotes training solely on synthesized target-object trajectories. Best values in bold.}
\label{tab:app_bimanual_object_adaptation}
\begin{tabular*}{\columnwidth}{@{\extracolsep{\fill}}lccccc@{}}
\toprule
\textbf{Training data}&\shortstack{\textbf{Pick Two}\\\textbf{Bottles}}&\shortstack{\textbf{Lift}\\\textbf{Pot}}&\shortstack{\textbf{Handover}\\\textbf{Microphone}}&\shortstack{\textbf{Put Bread}\\\textbf{into Pan}}&\textbf{Avg.}\\
\midrule
RoboTwin-50 &28&14&11&20&18.25\\
Replace-50 &41&46&23&35&36.25\\
Replace-100 &\textbf{48}&\textbf{60}&\textbf{40}&\textbf{43}&\textbf{47.75}\\
\bottomrule
\end{tabular*}
\end{table}

%% file: floats/appendix/bimanual_object_examples.tex
\begin{figure}[t]
\centering
\begingroup
\definecolor{bimanualheader}{HTML}{FFE99F}
\setlength{\fboxsep}{0pt}
\newcommand{\bimanualheading}[1]{\colorbox{bimanualheader}{\parbox[c][19pt][c]{0.247\columnwidth}{\centering\sffamily\bfseries\fontsize{7}{8}\selectfont #1}}}
\newcommand{\bimanualpanel}[1]{\includegraphics[width=0.247\columnwidth]{figures/appendix/bimanual_object_adaptation/#1}}
\noindent\bimanualheading{Pick Two\\Bottles}\hfill\bimanualheading{Lift Pot}\hfill\bimanualheading{Handover\\Microphone}\hfill\bimanualheading{Put Bread\\into Pan}\par
\vspace{0.8pt}
\noindent\bimanualpanel{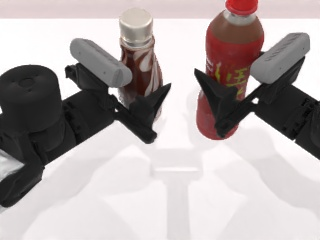}\hfill\bimanualpanel{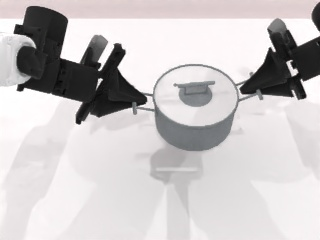}\hfill\bimanualpanel{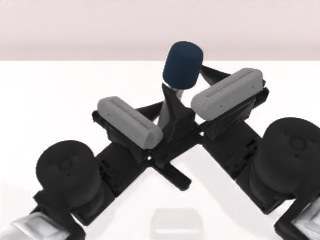}\hfill\bimanualpanel{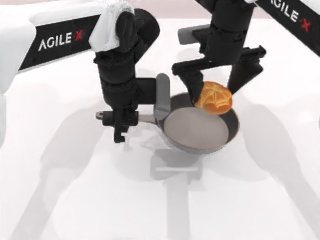}\par
\vspace{0.8pt}
\noindent\bimanualpanel{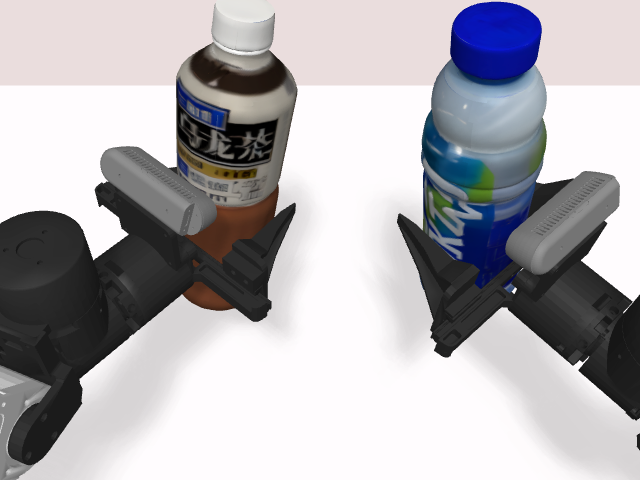}\hfill\bimanualpanel{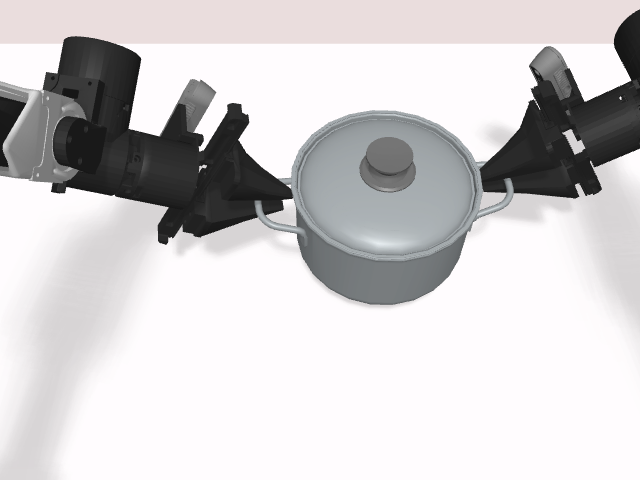}\hfill\bimanualpanel{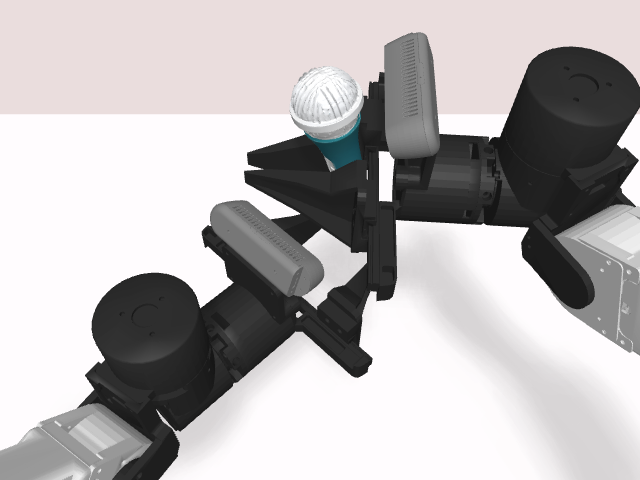}\hfill\bimanualpanel{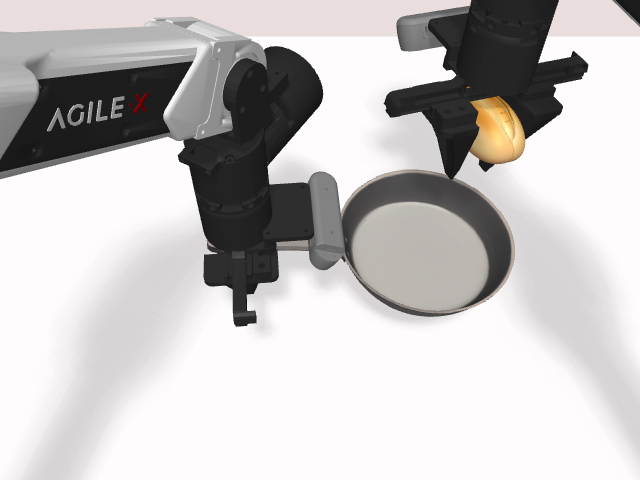}
\endgroup
\caption{\textbf{Bimanual object adaptation examples.} Top: source demonstrations in SAPIEN~\cite{xiang2020sapien}. Bottom: corresponding adaptations to new assets in MuJoCo~\cite{todorov2012mujoco}.}
\label{fig:app_bimanual_object_examples}
\end{figure}

%% file: floats/appendix/visual_alignment_metrics.tex
\begin{table}[t]
\centering\footnotesize
\setlength{\tabcolsep}{1.5pt}
\renewcommand{\arraystretch}{1.02}
\setlength{\aboverulesep}{2pt}
\setlength{\belowrulesep}{2pt}
\caption{\textbf{Reconstruction quality on DROID.} Mean full-image metrics over 20 initial scenes using RoboSnap\cite{zhang2026robosnap}, AR2S\cite{chen2026agentic} and RoboRecast. Best values in bold.}
\label{tab:app_visual_alignment}
\begin{tabular*}{\columnwidth}{@{\extracolsep{\fill}}lrrrrr@{}}
\toprule
Method & PSNR$\uparrow$ & SSIM$\uparrow$ & LPIPS$\downarrow$ & SIFT-MR$\uparrow$ & RGB-EMD$\downarrow$\\
\midrule
RoboSnap &13.60&0.602&0.400&0.167&22.28\\
AR2S &13.48&0.632&0.339&0.197&25.39\\
Ours &\textbf{14.97}&\textbf{0.675}&\textbf{0.293}&\textbf{0.204}&\textbf{17.82}\\
\bottomrule
\end{tabular*}
\end{table}

%% file: floats/appendix/single_arm_augmentation.tex
\begin{table*}[t]
\centering\small
\setlength{\tabcolsep}{3pt}
\renewcommand{\arraystretch}{1.02}
\setlength{\aboverulesep}{2pt}
\setlength{\belowrulesep}{2pt}
\caption{\textbf{Single-arm demonstration augmentation.} Generation success (\%) and joint smoothness under layout changes (top) and robot transfer (bottom). Best values within each comparison are bold.}
\label{tab:app_single_arm_augmentation}
\begin{tabular*}{\textwidth}{@{\extracolsep{\fill}}ll*{6}{c}@{}}
\toprule
\multirow{2}{*}{\textbf{Task / Target robot}} & \multirow{2}{*}{\textbf{Method}} & \multicolumn{4}{c}{\textbf{Generation success (\%)} $\uparrow$} & \multirow{2}{*}{\textbf{J-LDLJ} $\uparrow$} & \multirow{2}{*}{\textbf{J-SPARC} $\uparrow$} \\
\cmidrule(lr){3-6}
 & & D0 & D1 & D2 & Overall & & \\
\midrule
\multicolumn{8}{l}{\textit{Layout augmentation}} \\
\multirow{3}{*}{Square} & MimicGen~\cite{mandlekar2023mimicgen} & 20.0 & 10.0 & 0.0 & 10.0 & $-$15.92 & $-$4.39 \\
 & SkillGen~\cite{garrett2024skillmimicgen} & 20.0 & 40.0 & 15.0 & 25.0 & $-$16.63 & $-$4.03 \\
 & Ours & \textbf{100.0} & \textbf{100.0} & \textbf{100.0} & \textbf{100.0} & \textbf{\boldmath$-$15.48} & \textbf{\boldmath$-$3.74} \\
\cmidrule(lr){1-8}
\multirow{3}{*}{Assembly} & MimicGen & 35.0 & 40.0 & 30.0 & 35.0 & $-$18.83 & $-$5.23 \\
 & SkillGen & 30.0 & 65.0 & 20.0 & 38.3 & $-$18.54 & $-$4.88 \\
 & Ours & \textbf{100.0} & \textbf{100.0} & \textbf{100.0} & \textbf{100.0} & \textbf{\boldmath$-$17.88} & \textbf{\boldmath$-$4.14} \\
\cmidrule(lr){1-8}
\multirow{3}{*}{Coffee} & MimicGen & 85.0 & 85.0 & 20.0 & 63.3 & $-$17.75 & $-$4.82 \\
 & SkillGen & 80.0 & 85.0 & 45.0 & 70.0 & $-$18.03 & $-$4.71 \\
 & Ours & \textbf{100.0} & \textbf{100.0} & \textbf{100.0} & \textbf{100.0} & \textbf{\boldmath$-$15.93} & \textbf{\boldmath$-$3.70} \\
\midrule
\multicolumn{8}{l}{\textit{Robot augmentation}} \\
\multirow{3}{*}{Sawyer} & MimicGen~\cite{mandlekar2023mimicgen} & 25.0 & 15.0 & 0.0 & 13.3 & $-$17.26 & $-$5.07 \\
 & SkillGen~\cite{garrett2024skillmimicgen} & 30.0 & 25.0 & 10.0 & 21.7 & $-$18.20 & $-$4.95 \\
 & Ours & \textbf{100.0} & \textbf{100.0} & \textbf{100.0} & \textbf{100.0} & \textbf{\boldmath$-$17.12} & \textbf{\boldmath$-$4.58} \\
\cmidrule(lr){1-8}
\multirow{3}{*}{IIWA} & MimicGen & 45.0 & 55.0 & 25.0 & 41.7 & $-$16.75 & $-$5.11 \\
 & SkillGen & 50.0 & 70.0 & 50.0 & 56.7 & $-$18.09 & $-$5.33 \\
 & Ours & \textbf{100.0} & \textbf{100.0} & \textbf{100.0} & \textbf{100.0} & \textbf{\boldmath$-$16.72} & \textbf{\boldmath$-$4.72} \\
\cmidrule(lr){1-8}
\multirow{3}{*}{UR5e} & MimicGen & 80.0 & 80.0 & 25.0 & 61.7 & $-$16.94 & $-$5.39 \\
 & SkillGen & 75.0 & 80.0 & 85.0 & 80.0 & $-$18.29 & $-$5.43 \\
 & Ours & \textbf{100.0} & \textbf{100.0} & \textbf{100.0} & \textbf{100.0} & \textbf{\boldmath$-$16.50} & \textbf{\boldmath$-$5.22} \\
\bottomrule
\end{tabular*}
\end{table*}

%% file: floats/appendix/object_augmentation.tex
\begin{table}[t]
\centering\footnotesize
\setlength{\tabcolsep}{2pt}
\renewcommand{\arraystretch}{1.02}
\setlength{\aboverulesep}{2pt}
\setlength{\belowrulesep}{2pt}
\caption{\textbf{Object augmentation on Mug Cleanup.} Generation success and joint smoothness across three asset settings. Best values within each setting are bold.}
\label{tab:app_object_augmentation}
\begin{tabular*}{\columnwidth}{@{\extracolsep{\fill}}llccc@{}}
\toprule
\textbf{Setting} & \textbf{Method} & \shortstack{\textbf{Success}\\(\%) $\uparrow$} & \textbf{J-LDLJ} $\uparrow$ & \textbf{J-SPARC} $\uparrow$ \\
\midrule
\multirow{3}{*}{D0} & MimicGen~\cite{mandlekar2023mimicgen} & 36.7 & $-$19.55 & $-$5.37 \\
 & SkillGen~\cite{garrett2024skillmimicgen} & 40.0 & $-$20.16 & $-$5.57 \\
 & Ours & \textbf{100.0} & \textbf{\boldmath$-$17.88} & \textbf{\boldmath$-$5.31} \\
\cmidrule(lr){1-5}
\multirow{3}{*}{O1} & MimicGen & 43.3 & $-$19.55 & $-$5.71 \\
 & SkillGen & 43.3 & $-$20.26 & $-$5.66 \\
 & Ours & \textbf{100.0} & \textbf{\boldmath$-$17.79} & \textbf{\boldmath$-$5.00} \\
\cmidrule(lr){1-5}
\multirow{3}{*}{O2} & MimicGen & 26.7 & $-$19.98 & $-$5.85 \\
 & SkillGen & 38.3 & $-$20.15 & $-$5.94 \\
 & Ours & \textbf{100.0} & \textbf{\boldmath$-$17.50} & \textbf{\boldmath$-$5.02} \\
\bottomrule
\end{tabular*}
\end{table}

%% file: floats/appendix/bimanual_layout_augmentation.tex
\begin{table*}[t]
\centering\small
\setlength{\tabcolsep}{3pt}
\renewcommand{\arraystretch}{1.02}
\setlength{\aboverulesep}{2pt}
\setlength{\belowrulesep}{2pt}
\caption{\textbf{Bimanual layout augmentation.} Generation success (\%) and joint smoothness across three layout distributions. Best values within each task are bold.}
\label{tab:app_bimanual_layout_augmentation}
\begin{tabular*}{\textwidth}{@{\extracolsep{\fill}}ll*{6}{c}@{}}
\toprule
\multirow{2}{*}{\textbf{Task}} & \multirow{2}{*}{\textbf{Method}} & \multicolumn{4}{c}{\textbf{Generation success (\%)} $\uparrow$} & \multirow{2}{*}{\textbf{J-LDLJ} $\uparrow$} & \multirow{2}{*}{\textbf{J-SPARC} $\uparrow$} \\
\cmidrule(lr){3-6}
 & & P0 & P1 & P2 & Overall & & \\
\midrule
\multirow{2}{*}{Threading} & DexMimicGen~\cite{jiang2025dexmimicgen} & 45.0 & 20.0 & 70.0 & 45.0 & $-$18.45 & $-$4.86 \\
 & Ours & \textbf{100.0} & \textbf{100.0} & \textbf{100.0} & \textbf{100.0} & \textbf{\boldmath$-$16.20} & \textbf{\boldmath$-$3.68} \\
\cmidrule(lr){1-8}
\multirow{2}{*}{Assembly} & DexMimicGen & 15.0 & 25.0 & 30.0 & 23.3 & $-$19.57 & $-$6.81 \\
 & Ours & \textbf{100.0} & \textbf{100.0} & \textbf{100.0} & \textbf{100.0} & \textbf{\boldmath$-$16.31} & \textbf{\boldmath$-$4.67} \\
\cmidrule(lr){1-8}
\multirow{2}{*}{Transport} & DexMimicGen & 25.0 & 10.0 & 0.0 & 11.7 & $-$20.10 & $-$6.11 \\
 & Ours & \textbf{100.0} & \textbf{100.0} & \textbf{100.0} & \textbf{100.0} & \textbf{\boldmath$-$18.68} & \textbf{\boldmath$-$6.01} \\
\bottomrule
\end{tabular*}
\end{table*}

%% file: floats/visual_alignment.tex
\begin{figure}[t]
\centering
\begingroup
\definecolor{alignmentheader}{HTML}{FFE99F}
\setlength{\fboxsep}{0pt}
\newcommand{\alignmentheading}[1]{\colorbox{alignmentheader}{\parbox[c][19pt][c]{0.2485\columnwidth}{\centering\sffamily\bfseries\fontsize{7}{8}\selectfont #1}}}
\noindent\alignmentheading{Real\\observation}\hfill\alignmentheading{RoboSnap}\hfill\alignmentheading{AR2S}\hfill\alignmentheading{Ours}\par
\vspace{0.5pt}
\noindent\includegraphics[width=\columnwidth,trim=0 0 0 66bp,clip]{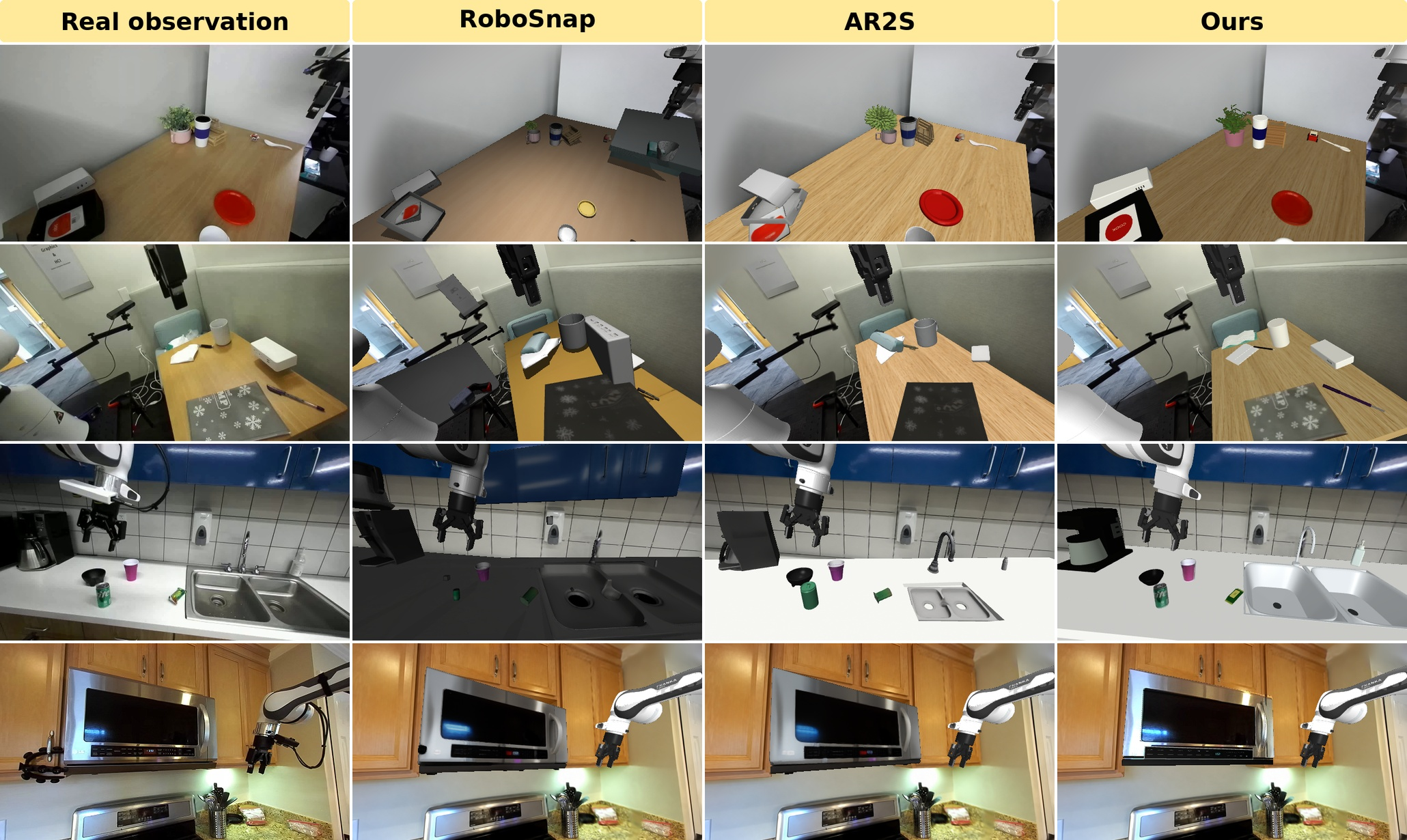}
\endgroup
\caption{\textbf{Visual alignment on DROID.} Initial reconstructions of four representative scenes, with the same InfiniSplat~\cite{wang2026infinisplat} background across methods.}
\label{fig:visual_alignment}
\end{figure}

%% file: floats/appendix/augmentation_examples.tex
\begin{figure}[t]
\centering
\includegraphics[width=\columnwidth]{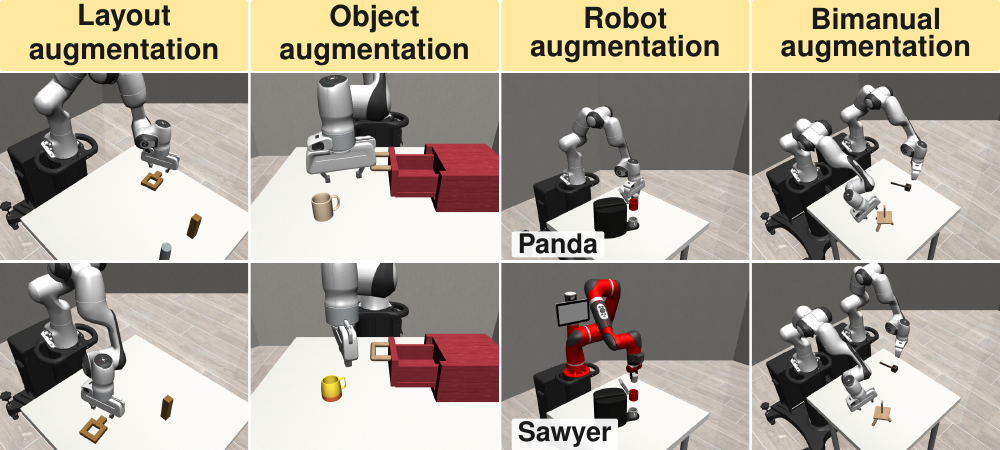}
\caption{\textbf{Demonstration augmentation examples.} Top: source demonstrations. Bottom: RoboRecast adaptations to new layouts, objects, and robots, including bimanual manipulation.}
\label{fig:app_augmentation_examples}
\end{figure}

%% file: floats/appendix/generation_cost.tex
\begin{table}[t]
\centering\footnotesize
\setlength{\tabcolsep}{2pt}
\renewcommand{\arraystretch}{1.02}
\setlength{\aboverulesep}{2pt}
\setlength{\belowrulesep}{2pt}
\caption{\textbf{Generation time and cost across agent models.} Mean times and API-equivalent costs over five tasks. Best in bold.}
\label{tab:app_generation_cost}
\begin{tabular*}{\columnwidth}{@{\extracolsep{\fill}}lcccc@{}}
\toprule
\multirow{2}{*}{\textbf{Model}} & \multicolumn{3}{c}{\textbf{Time (min)} $\downarrow$} & \multirow{2}{*}{\textbf{USD} $\downarrow$} \\
\cmidrule(lr){2-4}
 & Init. & Opt. & Total & \\
\midrule
GPT-5.6-Sol & 35.77 & 47.76 & 83.53 & 23.32 \\
GPT-6.1-Sol & 16.65 & 12.36 & 29.01 & \textbf{3.10} \\
GPT-6-Astra & \textbf{14.89} & \textbf{11.95} & \textbf{26.84} & 21.52 \\
\bottomrule
\end{tabular*}
\end{table}

%% file: sec/appendix/c_limitations_and_future_work.tex
\section{Limitations and Future Work}
\label{sec:limitations_future_work}

Further improvements to RoboRecast concern generation efficiency, support for mobile manipulation, and the visual realism and diversity of synthesized observations.

\phantomsection\label{sec:experience_library}
\noindent\textbf{Generation Efficiency and Experience Reuse.} Although model choice affects generation cost, processing each demonstration still requires scene modeling and repeated simulation to diagnose and correct interaction failures.
Similar failures can recur across demonstrations, motivating the reuse of previously effective repairs.
As a preliminary exploration, our Experience Library stores resolved failures together with their diagnostic evidence and the modeling or motion edits that addressed them.
The agent retrieves these records to guide subsequent revisions and verifies their applicability through simulation in the current scene.

During real-to-sim reconstruction of demonstrations we collected on physical UR5 for four tasks, we build the library from the first 40 demonstrations per task and reconstruct the remaining ten with and without reuse.
Experience reuse reduces mean end-to-end generation time by 4.9\%--21.0\% across the four tasks.
These preliminary results suggest that repairs learned from earlier demonstrations can reduce repeated trial and error within the same task.
Extending this benefit across tasks will require identifying which aspects of a repair generalize and when they remain applicable to new objects and interactions.

\input{floats/mobile_manipulation_demo}
\noindent\textbf{Mobile Manipulation.} Our policy evaluations focus on fixed-base single-arm and bimanual manipulation.
Figure~\ref{fig:mobile_manipulation_demo} illustrates a preliminary extension to mobile manipulation, showing a real chair-pushing demonstration and its simulated counterpart.
This setting requires coordinating base and arm motion to maintain contact with the chair as the robot moves.
Future work could extend diagnosis-driven optimization to jointly refine base trajectories and arm motions, and evaluate the resulting demonstrations through mobile manipulation policy learning.

\noindent\textbf{Visual Diversity and Realism.} Even when simulated rollouts satisfy the interaction requirements, their rendered observations can differ from real images in appearance and lack visual diversity.
Future work could use video generation and editing conditioned on simulated rollouts to improve visual realism and diversify lighting, backgrounds, and task-irrelevant objects.
Such edits should preserve object motion and contact timing so that the enriched observations remain aligned with robot actions and retain the physical and semantic validity of the underlying interactions.

%% file: floats/mobile_manipulation_demo.tex
\begin{figure}[!t]
  \centering
  \includegraphics[width=\columnwidth]{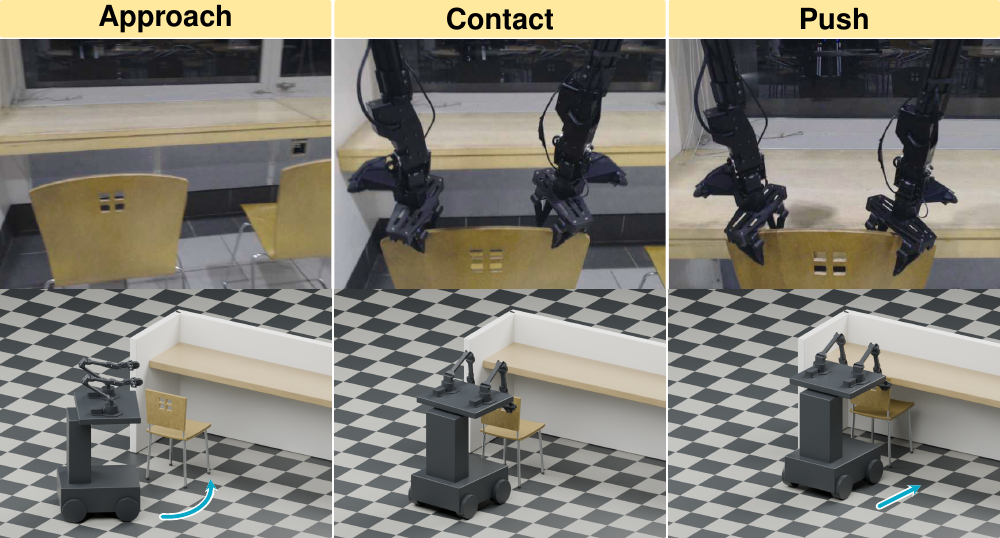}
  \caption{\textbf{Mobile chair pushing.} Real (top) and simulated (bottom) sequences show approach, contact, and push.}
  \label{fig:mobile_manipulation_demo}
\end{figure}